\documentclass[conference]{IEEEtran}
\IEEEoverridecommandlockouts
\usepackage{dsfont}
\usepackage{cite}
\usepackage{amsmath,amssymb,amsfonts,bm}
\usepackage{graphicx}
\usepackage{textcomp}
\usepackage{xcolor}
\usepackage{mathtools}
\usepackage{placeins}

\usepackage{dsfont}
\usepackage{cite}
\usepackage{amsmath,amssymb,amsfonts,bm}
\usepackage{graphicx}
\usepackage{textcomp}
\usepackage{xcolor}
\usepackage{mathtools}
\usepackage{placeins}
\usepackage[caption=false,font=normalsize,labelfont=sf,textfont=sf]{subfig}
\usepackage{enumitem}

\usepackage{bbm}

\usepackage{algorithm,algpseudocode}
\usepackage{hyperref}

\usepackage{algorithm,algpseudocode,amsmath}
\def\BibTeX{{\rm B\kern-.05em{\sc i\kern-.025em b}\kern-.08em
    T\kern-.1667em\lower.7ex\hbox{E}\kern-.125emX}}
    
\begin{document}



\title{Federated Learning-based Collaborative Wideband Spectrum Sensing and Scheduling for UAVs in UTM Systems} 
    \author{
        \IEEEauthorblockN{Sravan Reddy Chintareddy\IEEEauthorrefmark{1}, Keenan Roach\IEEEauthorrefmark{2}, Kenny Cheung\IEEEauthorrefmark{2}, Morteza Hashemi\IEEEauthorrefmark{1}}
        \IEEEauthorblockA{
        \IEEEauthorrefmark{1}Department of Electrical Engineering and Computer Science, University of Kansas \\
    \IEEEauthorrefmark{2}Universities Space Research Association (USRA)
        }
    }


\maketitle


\begin{abstract}

    In this paper, we propose a data-driven framework for collaborative wideband spectrum sensing and scheduling for networked unmanned aerial vehicles (UAVs), which act as the secondary users (SUs) to opportunistically utilize detected ``spectrum holes''. Our overall framework consists of three main stages. Firstly, in the \emph{model training} stage, we explore dataset generation in a multi-cell environment and training a machine learning (ML) model using the federated learning (FL) architecture. Unlike the existing studies on FL for wireless that presume datasets are readily available for training, we propose a novel architecture that directly integrates wireless dataset generation, which involves capturing I/Q samples from over-the-air signals in a multi-cell environment, into the FL training process. To this purpose, we propose a multi-label classification problem for wideband spectrum sensing to detect multiple spectrum holes simultaneously based on the I/Q samples collected locally by the UAVs. In the traditional FL that employ FedAvg as the aggregating method, each UAV is assigned an equal weight during model aggregation. However, due to the disparities in channel conditions in a multi-cell environment, the FedAvg approach may not generalize effectively for all the UAV locations. To address this issue, we propose a proportional weighted federated averaging method (pwFedAvg) in which the aggregating weights incorporate wireless channel conditions and received signal powers at each individual UAV. As such, the proposed method integrates the intrinsic properties of wireless datasets into the FL algorithm.   
    Secondly, in the \emph{collaborative spectrum inference} stage, we propose a collaborative spectrum fusion strategy that is compatible with the unmanned aircraft system traffic management (UTM) ecosystem.
    In particular, we improve the accuracy of spectrum sensing results by combining the individual multi-label classification results from the individual UAVs at a central server. Finally, in the \emph{spectrum scheduling} stage, we leverage reinforcement learning (RL) solutions to dynamically allocate the detected spectrum holes to the secondary users. To evaluate the proposed methods, we establish a comprehensive simulation framework that generates a near-realistic synthetic dataset using MATLAB LTE toolbox by incorporating base-station~(BS) locations in a chosen area of interest, performing ray-tracing, and emulating the primary users channel usage in terms of I/Q samples. This evaluation methodology provides a flexible framework  to generate \emph{large spectrum datasets} that could be used for developing ML/AI-based spectrum management solutions for aerial devices.
\end{abstract}
\maketitle
\section{Introduction}
\label{sec:intro}
\begin{figure}[t]
    \centering \includegraphics[width=.8\linewidth]{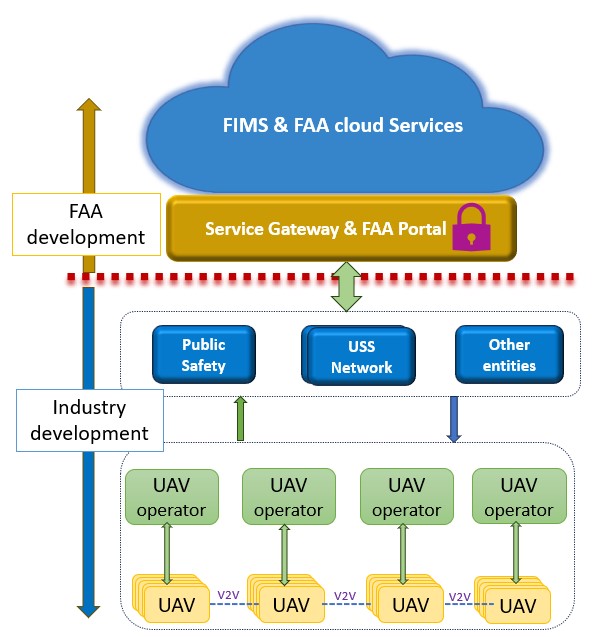}
  \caption{Unmanned Aircraft System Traffic Management (UTM) architecture showing the separation between Federal Aviation Administration (FAA) and industry developments; Flight Information Management System (FIMS).}
  \label{fig:utmmodel}
\end{figure}
Unmanned aerial vehicles (UAVs) have attracted significant interest from communications and networking, robotics, and control societies for exploring novel applications such as on-demand connectivity, search-and-rescue operations, situational awareness, to name a few~\cite{menouar2017uav}. As of April 2024, there were roughly 800,000 registered UAVs in the US alone, positioning UAVs as one of the fastest-growing sectors in the aviation industry~\cite{dronenumbers}. Traditionally, UAVs that are used for recreational purposes are operated under visual line of sight (VLOS) conditions. However, real-world and commercial deployments will most likely be in the form of beyond visual line-of-sight (BVLOS), which provides easier access to remote or hazardous areas, less human intervention, and reduced cost of operation~\cite{li2019beyond}. For safe operations of multiple UAVs under BVLOS conditions, NASA and FAA are in the process of defining the UTM system~\cite{kopardekar2016unmanned}. Fig.~\ref{fig:utmmodel} shows a simplified form of the UTM architecture, highlighting the separation between FAA and industry development and deployment responsibilities for the necessary infrastructure, services, and entities that interact within the UTM ecosystem. In this work, we mainly focus on the hierarchical structure between multiple operators and the UAS service supplier (USS), which assists multiple operators in meeting UTM operational requirements, ensuring safe and efficient utilization of the airspace.

The concept of operations within the UTM architecture~\cite{kopardekar2016unmanned} highlights the need for spectrum resources to facilitate wireless communications between UAVs,  UAV operators, and the USS network.
Existing terrestrial mobile networks (for example, 4G LTE and the upcoming 5G-and-beyond) provide significant wireless coverage with relatively low latency, high throughput, and low cost, making the cellular network a good candidate for the operation of UAVs in BVLOS scenarios~\cite{abdalla2021communications}. However, the proliferation of new wireless services and the demand for higher cellular data rates have significantly exacerbated the spectrum crunch that cellular providers are already experiencing. 
Therefore, it is essential to develop dynamic spectrum sensing, inference, and sharing solutions for UAV operations in existing licensed and unlicensed spectrum to enable advanced aerial use cases in BVLOS, such as urban air mobility (UAM) and advanced air mobility (AAM)~\cite{rimjha2021urban,ghazikor2023exploring}. 


\begin{figure}[t]
    \centering
    \includegraphics[width=0.8\linewidth]{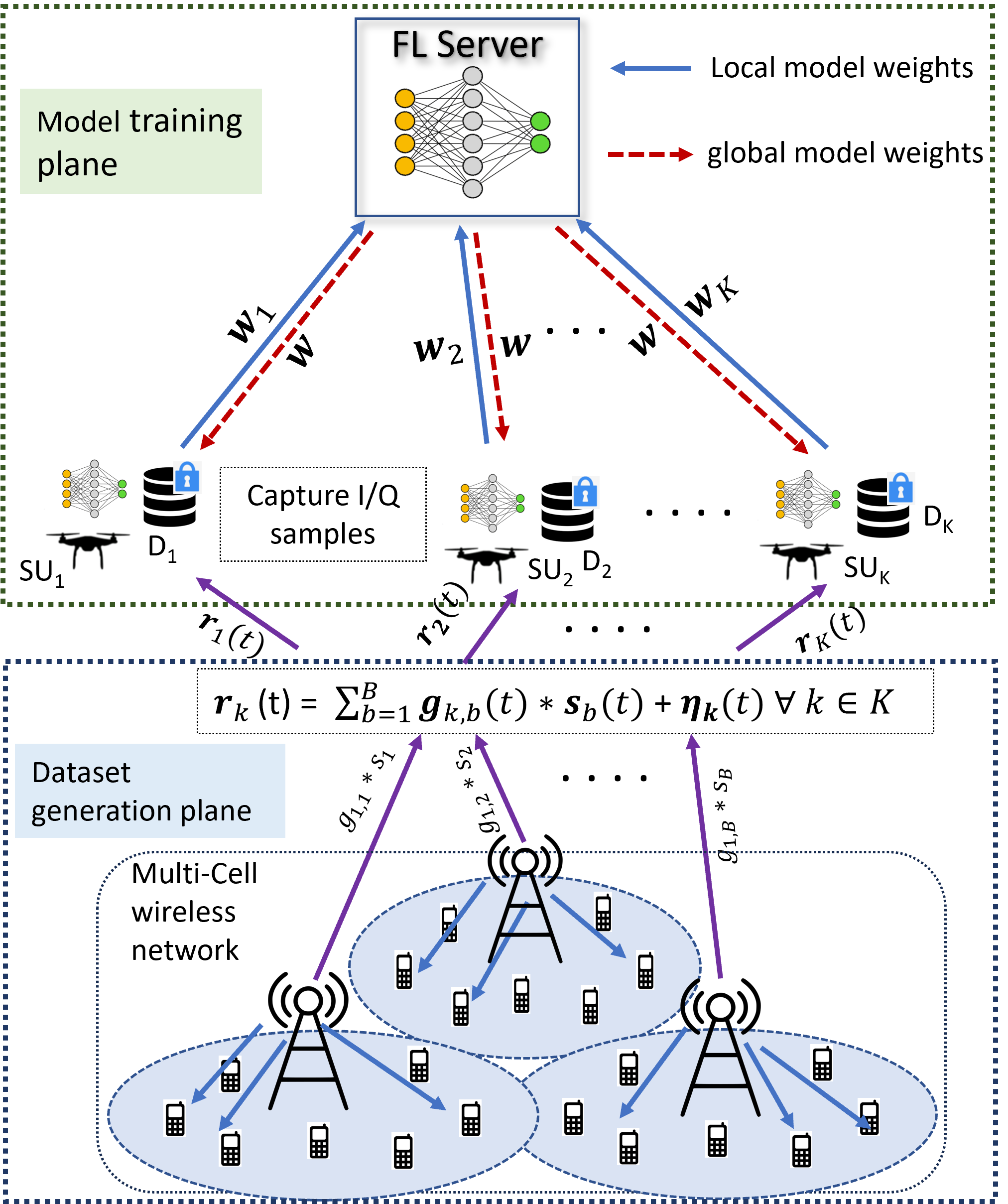}
    \caption{Envisioned FL system model in a Multi-cell wireless network with multiple UAVs.}
    \label{fig:FL-sysmodel}
\end{figure}


There exists a multitude of prior works on spectrum management frameworks for ground users~\cite{ahmad20205g,uvaydov2021deepsense,cui2019multi,li2020deep,nguyen2018deep}. For instance, the authors in~\cite{li2020deep,uvaydov2021deepsense,cui2019multi} propose deep learning-based wideband spectrum sensing to dynamically detect ``spectrum holes''. Furthermore, the authors in~\cite{nguyen2018deep} propose reinforcement learning~(RL) techniques for spectrum sharing, assuming that spectrum sensing results are readily available. While these data-driven spectrum management frameworks for ground users are available, they are not directly applicable for UTM-enabled UAV operations, due to several factors, such as the widely different wireless channel models and the overall system architecture~\cite{uvaydov2021deepsense,nguyen2018deep}. 
In the context of UAV spectrum sharing systems, the authors in~\cite{kakar2017waveform,shang2020spectrum} proposed spatial spectral sensing~(SSS) to develop efficient spectrum sharing policies for UAV communications aimed at improving the overall spectral efficiency~(SE). However, the SSS models do not consider the spectrum usage pattern of users under realistic scenarios (e.g., ignoring the I/Q level samples), and/or they consider only a single primary user (PU) or secondary user (SU). Moreover, the problem of joint multi-channel wideband spectrum sensing and scheduling among several SUs has not been fully investigated. 
In this paper, we propose a unified and data-driven spectrum sensing and scheduling framework to enable UAVs to effectively share the spectrum with existing primary users. 
To make our development more concrete and grounded, the problem of joint spectrum sensing and sharing is formulated as an energy efficiency (EE) maximization in a wideband multi-UAV network scenario. Then, we transform the EE optimization problem into a Markov Decision Process (MDP) to maximize the overall throughput of the SUs. At the spectrum sensing stage, we note the inherent hierarchical nature of the UTM architecture with USS (shown in Fig.~\ref{fig:utmmodel}) is a good match for federated learning (FL) based spectrum sensing. For spectrum scheduling stage, we develop RL-based solutions to enable non-manual and automated spectrum resource allocation. 
Particular to the spectrum sensing stage, we propose an FL-based cooperative wideband spectrum sensing across multiple UAVs. To this purpose, we develop a multi-label classification framework to identify spectrum holes based on the observed I/Q samples. Each UAV trains their respective local models using the locally collected datasets and transmits the local model parameters to the central server. Furthermore, we propose a novel proportional
weighted federated averaging~(pwFedAvg) method that incorporates the power level received at each UAV into the FL aggregation algorithm, thereby integrating the dataset generation plane with the FL model training plane, as shown in Fig.~\ref{fig:FL-sysmodel}. 
Once the training process is completed, all UAVs have an updated global model that predicts \emph{spectrum holes}. 
To further enhance the accuracy of the individual spectrum inference results, the predicted spectrum holes from the multi-label classification at each UAV are fused at a central server within the UTM ecosystem. In the spectrum scheduling stage, we develop and implement several RL algorithms, including the standard Q-learning methods to dynamically allocate underutilized spectrum sub-channels to multiple UAVs. We further investigate the performance of the ``vanilla'' deep Q-Network (DQN) and its variations, including double DQN (DDQN) and DDQN with soft-update.

Furthermore, one of the primary challenges of using machine learning (ML) based methods for spectrum sensing and scheduling approaches is the need for large amounts of training data. The lack of available spectral data in many cases is a significant obstacle, especially for UAV networks that introduce an additional level of complexity for large-scale experimental data collection. To address this gap, we have developed a comprehensive framework for generating spectrum datasets. This framework models LTE waveform generation and propagation channel in any environment of interest, particularly suitable for UTM-enabled UAV applications. Using the generated dataset, we provide a comprehensive set of numerical results to demonstrate the efficacy of the joint FL-based spectrum sensing, spectrum fusion and RL-based dynamic spectrum allocation to multiple UAVs. 
In summary, the main contributions of this paper are as follows:
\begin{itemize}
    \item We develop a spectrum management framework based on the envisioned UTM deployment architecture. To this end, we propose a joint spectrum sensing and scheduling problem for collaborative networked UAVs that operate according to the UTM rules. The joint optimization problem integrates the spectrum sensing results into the spectrum scheduling stage for scenarios with multiple secondary users (i.e., UAVs) and primary users. 
    \item For spectrum sensing, we propose an FL-based solution to enable collaborative model training across distributed UAVs.  We propose the pwFedAvg method that integrates the underlying wireless channel conditions into the FL aggregation step. We also provide the convergence analysis results of the proposed pwFedAvg method.  We demonstrate the benefits of collaborative spectrum sensing through a fusion step. For the spectrum scheduling stage, we develop RL-based solutions leveraging DQN-based approaches.
    \item  We outline a methodology for generating large amounts of I/Q dataset for UAVs in a wide geographical area, considering the effects in a multi-cell multi-path environment by incorporating the base-station locations and accurately modeling the environment using ray-tracing methods. 
Based on the established framework, we provide a comprehensive set of numerical results to analyze the performance of pwFedAvg compared with the traditional FedAvg approach, as well as with centralized and local learning. Our results demonstrate the efficacy of the pwFedAvg method for collaborative spectrum sensing, without the need to transfer all I/Q samples to one location as in central learning. 
\end{itemize}


\noindent 
This paper extends our prior work~\cite{chintareddy2023collaborative} in which we did not investigate the feasibility of  model training using FL methods for UAVs. In contrast, this paper mainly focuses on developing FL-based spectrum sensing by incorporating the wireless datasets captured by multiple UAVs into the FL model training plane, as shown in Fig.~\ref{fig:FL-sysmodel}. Furthermore, we have significantly extended our dataset generation by scaling the size of captured I/Q data samples and increasing the number of reflection and diffraction rays, thereby enhancing the fidelity of emulating the propagation environment.  
The remainder of this paper is organized as follows. In Section~\ref{sec:relworks}, we review related works. In Section~\ref{sec:sysmodel}, we present the overall system model and problem formulation for FL-based wideband spectrum sensing and collaborative spectrum inference and scheduling. In Section~\ref{sec:proposedflsoln}, we discuss the dataset generation model and the model training aspects of FL based solution to incorporate our proposed pwFedAvg method, followed by a discussion of the convergence analysis of pwFedAvg. In Section~\ref{sec:DSARL}, we present dynamic spectrum allocation using RL. Section~\ref{sec:datasetgen} describes our methodology to generate synthetic spectrum dataset followed by our numerical results in Section~\ref{sec:results}. Finally, Section~\ref{sec:conclusion} concludes the paper.

\section{Related Works}
\label{sec:relworks}
\subsection{Spectrum Sensing and Sharing for UAVs} 
 
The authors in~\cite{shang2019spatial} address spectrum access and interference management by utilizing SSS for ground based device-to-device (D2D) communications~\cite{chen2016spatial,chen2018qos}. Furthermore, the authors in~\cite{shang2020spectrum, shang20203d} extend the usage of SSS to UAVs to opportunistically access
the licensed channels that are occupied by the D2D communications of ground users. The UAVs perform SSS to obtain the received signal strength and compare it with a threshold to identify the spectrum occupancy of a particular D2D channel. 
However, in general, energy-based detection methods would require capturing the entire waveform for a sub-channel to compute the energy and compare it with a predefined threshold. When there are multiple sub-channels, such detection methods repeated for each sub-channel add further time and hardware complexity. Therefore, SSS methods are not directly applicable to wideband spectrum sensing by UAVs to detect multiple spectrum holes \emph{simultaneously}.

In addition to the SSS methods, data-driven deep learning (DL) methods for spectrum sensing have been considered in prior works~\cite{liu2019deep,chew2020spectrum}. To develop multi-channel spectrum sensing using DL, the authors in~\cite{uvaydov2021deepsense} developed a fast wideband spectrum sensing based on DL. The DL model is based on a convolutional neural network (CNN) that accepts raw I/Q signals and predicts the spectrum holes. The above works consider a single PU, a single SU only, and the channel between the PU and SU is modeled as a Rayleigh fading channel.

Furthermore, there exists extensive research on spectrum sharing solutions. For example, the authors in~\cite{li2020deep,naparstek2017deep, naparstek2018deep, albinsaid2021multi} propose the use of RL for dynamic spectrum access in multi-channel wireless networks. Furthermore, the authors in ~\cite{nguyen2018deep,bokobza2023deep,wang2018deep} propose the use of DQN, where, in each time slot, a single SU decides whether to stay idle or transmit using one of the sub-channels in a multi-channel environment without performing spectrum sensing. While these studies have provided significant insights, they consider one SU only and are well studied for ground based communications.

In this paper, we consider a data-driven approach to predict multiple spectrum holes \emph{simultaneously} from the raw I/Q signals captured in a multi-cell multi-path fading environment consisting of multiple PUs and SUs. We incorporate ray-tracing methods to effectively model the dynamic UAV environment instead of assuming a statistical channel model. Furthermore, we employ RL for dynamically allocating resources for the UAVs based on predicted spectrum holes.

\subsection{FL-based Spectrum Sensing}
FL for spectrum sensing has lately gained popularity~\cite{chen2021federated, gao2021fedswap,wasilewska2023secure,khalek2023advances,liu2022wireless}. The authors in~\cite{wasilewska2023secure,chen2021federated} discuss the application of FL for spectrum sensing in cognitive radio environments, where a SU detects the spectrum holes in the PU's spectrum band and utilize them opportunistically. However, the studies only considered a single PU with multiple SUs within the coverage of the PU. There exists a separate class of research that concentrate on interference management in a multi-cell wireless networks and incorporate over-the-air computation in FL~\cite{wang2022interference,shi2022multiple,xiao2023over}. For instance, the authors in~\cite{wang2022interference} study the adverse effects of inter-cell interference on the uplink and downlink local model aggregates and global model updates and propose solutions to mitigate the interference. In contrast to the above mentioned works, in this paper, we cater for multiple PUs, multiple SUs, incorporate co-channel interference at the dataset generation level, and also consider wideband spectrum sensing.


Additionally, there exists few research works on FL for wireless systems that investigate how the convergence of the learning process is affected by the noisy transmissions between the clients and the server~\cite{amiri2021convergence, wei2022federated}. However, these investigations often assume that the datasets are readily available at the clients and primarily focus on the specific discussions of model training plane, as illustrated in Fig.~\ref{fig:FL-sysmodel}. Furthermore, these studies consider standard ML datasets, such as CIFAR-10, MNIST, Shakespeare~\cite{wei2022federated}, while using traditional federated averaging algorithms. 

Yet, wireless datasets collected by multiple UAVs in a multi-cell environment are significantly complex and different compared to those standard datasets. For instance, data collected at one UAV location may encounter distinct wireless channels, varying numbers of propagation paths, and significantly different received signal power levels compared to the data collected at other locations. This variability underscores the need for tailored approaches to the model training within FL frameworks, particularly when dealing with datasets from real-world wireless environments. Hence, we propose a weighted averaging algorithm (pwFedAvg) that captures the disparities in the datasets captured at different UAV locations. Moreover, none of the previous works considered training the FL models with I/Q datasets. To address this gap, we propose a novel architecture to capture wireless datasets in a multi-cell environment, and integrate the dataset generation and the model training planes, to incorporate the effects of wireless datasets into the FL model training, as illustrated in Fig.~\ref{fig:FL-sysmodel}.

\section{System Model and Problem Formulation}
\label{sec:sysmodel}

\begin{figure}[t]
    \centering
    \includegraphics[width=\linewidth]{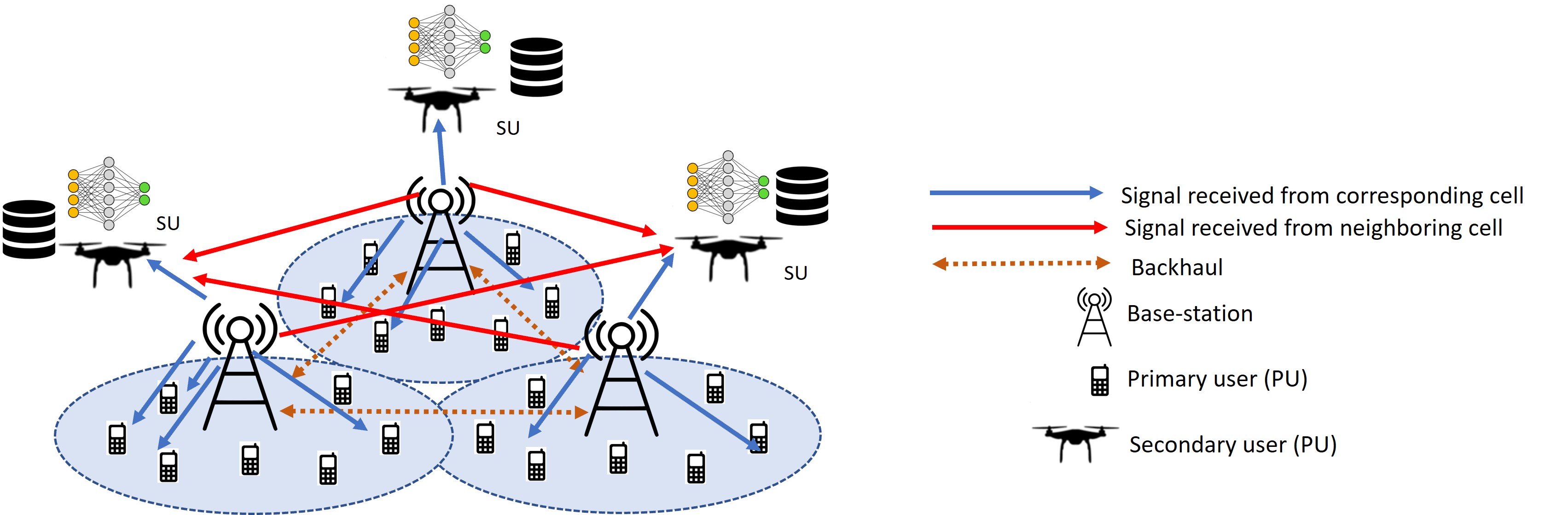}
    \vspace{-4mm}
    \caption{A zoomed in version of dataset generation plane of a Multi-cell wireless network with multiple UAVs.}
    \label{fig:ds-sysmodel}
    \vspace{-5mm}
\end{figure}

To model collaborative wideband spectrum sensing and scheduling, we consider a multi-cell wireless network that consists of a set of base-stations (BS) denoted by $\mathcal{B}$ ($|\mathcal{B}| = B$), as shown in Fig.~\ref{fig:ds-sysmodel}. In addition, we consider a set of UAVs denoted by $\mathcal{K}$ ($|\mathcal{K}| = K$) in the system. To coordinate the collaborative spectrum sensing, fusion, and scheduling, we assume that each time slot is divided into four consecutive sub-slots: UAV resource request~($t_{req}$), spectrum sensing~($t_s$), broadcasting to central server~($t_b$), and channel access~($t_a$).
Specifically, at the beginning of each time slot, the UAVs that require PU resources request the server for resource allocation. 
In the subsequent sub-slot of sensing~($t_s$), the UAVs perform spectrum sensing and broadcasts the sensed channel information in the following sub-slot~($t_b$). The central server then applies fusion rules and assigns spectrum holes to the requesting UAVs. The UAVs then transmit on the allocated spectrum holes in the access sub-slot~($t_a$). In this paper, we focus on three main stages to develop our proposed framework: (i) FL-based training for wideband spectrum sensing, (ii) collaborative spectrum inference and fusion, and (iii) spectrum scheduling. To coordinate the above three stages, we assume a central server within the UTM ecosystem. Next, we describe these stages.

\subsection{Model Training, Spectrum Inference, and Scheduling Stages}
\textbf{FL-based model training for spectrum sensing.} The UTM system architecture shown in Fig.~\ref{fig:utmmodel} supports data exchange between multiple UAVs through the USS network. Such a hierarchical architecture makes it feasible to implement FL-based learning algorithms to identify spectrum holes. In this case, we may consider two deployment models within the UTM architecture. One model would be to have a server deployed by each UAV operator where multiple UAVs connected to the operator act as FL clients. The second model would have a server within the USS network that orchestrates multiple UAV operators.
Thus, with several UAVs training local models, they exchange model parameters with the central server that is located either at the USS or UAV operator. The central server then aggregates the local model weights according to an aggregation algorithm and transmits the global model weights back to the UAVs to update their local models. 

\textbf{Collaborative spectrum inference and fusion.} Due to the highly dynamic environment in which UAVs operate, it may not be feasible for all UAVs to achieve high prediction accuracy across all sub-channels. Therefore, we leverage collaborative spectrum inference by the UAVs, and perform fusion at the fusion module within the central server to increase the reliability of spectrum hole detection.  In particular, 
each individual UAV captures the raw I/Q samples from over-the-air received signals and predicts the availability of spectrum holes across $M$ sub-channels using the FL-trained model. We assume that there is an associated spectrum inference cost for each UAV $k$ involved in sensing at time slot $t$. The spectrum inference cost is the energy consumed for sensing the spectrum and is proportional to the voltage $V_{CC}$ of the receiver, the system bandwidth ${W}$, and the duration allotted for sensing~($t_s$)~\cite{zhang2011mili}. Therefore, it is defined as  $SC_{k,m}(t) = t_sV_{CC}^2{{W}}_{m}$, where $W_m$  is the $m$-th sub-channel bandwidth. Upon completion of the spectrum inference phase, the UAV $k$
has a predicted spectrum occupancy vector $\bm{\widehat{h}}_k(t) = [{\widehat{h}}_{k, 1}(t), ..., {\widehat{h}}_{k,M}(t)]$ such that ${\widehat{h}}_{k, m}(t) = 0$ if the $m$-th sub-channel is detected vacant at time $t$, and ${\widehat{h}}_{k, m}(t) = 1$ otherwise. This problem can be considered as a multi-label classification problem, and we leverage deep neural network (DNN) at each UAV that accepts raw I/Q samples $\bm{R}_k$ as inputs and outputs the predicted spectrum occupancy vector $\bm{\widehat{h}}_k(t)$. 





The central server receives multiple copies of spectrum holes detected by individual UAVs and applies fusion rules that results in aggregated prediction. In this paper, we use the $n$-out-of-${K}$ fusion rule defined as follows: 

\vspace{-.2cm}
\begin{equation}
\label{eqn:fusion}
    z_m(t) = \begin{dcases}
    0,& \text{if } \sum_{k \in \mathcal{K}} \mathds{1} \{{\widehat{h}}_{k, m}(t)=0\}\geq n;\\
    1,              & \text{otherwise},  
\end{dcases}
\end{equation}
where $\mathds{1}\{.\}$ is an indicator function. In this case,  $\bm{z}(t)=[z_1(t), ..., z_M(t)]$ is the fused prediction of all the $M$ sub-channels at the central server. 
Note that when $n=1$, the $n$-out-of-${K}$ rule is equivalent to the ``OR'' rule, and $n={K}$ is the same as the ``AND'' rule.

 \textbf{Spectrum scheduling.}
\label{sec: access module}
Based on the aggregated fusion result provided by the fusion module, the central server then allocates sub-channels to the requesting UAVs. The UAVs then transmit data on the sub-channels allocated to them by the server in the next time step. 
The transmission energy consumption is denoted by  $AC_{k,m}(t)$. The access cost is the energy consumed for data transmission and is defined as $AC_{k,m}(t) = t_{a}P_{tx}$, 
where, $P_{tx}$ is the transmit power and $t_{a}$ is the time allotted to transmission. Furthermore, the transmission utility is the amount of data transmitted on the allocated sub-channel, which is defined as follows:
\begin{equation}
    U_{k,m}(t)= t_a {W}_m \log_2\left(1+\text{SNR}_{k,m}(t)\right),
\end{equation}
where $\text{SNR}_{k,m}(t)$ denotes the signal-to-noise ratio for UAV $k$ on sub-channel $m$. 

We highlight that the UAVs transmit on those sub-channels  that were detected vacant in the previous time step. Hence, spectrum collision occurs when the previously detected spectrum holes are no longer available at the current time step. We assume that the true state of sub-channel $m$ is denoted by $\bar{z}_m(t)$. To capture this, we define the spectrum access collision indicator $c_{k,m}(t)$ as follows: 
\begin{equation}
    c_{k,m}(t) = \begin{dcases}
        1, & \text{if } \bar{z}_{m}(t)=0~ \text{and}~  z_{m}(t-1) =0; \\
        -1, & \text{if } \bar{z}_{m} (t)\neq0~ \text{and} ~z_{m}(t - 1) =0; \\
        0, & \text{otherwise}. 
    \end{dcases}
\end{equation}
Next, we formulate a joint spectrum sensing and scheduling optimization problem.   
\subsection{Joint Spectrum Sensing and Scheduling Problem Formulation}
\label{sec:problem2}
Given the presented system model, we now introduce a joint spectrum sensing and scheduling problem to coordinate collaborative spectrum sensing and spectrum scheduling. We cast the problem as an EE optimization for the UAVs that opportunistically use the spectrum resources of the primary network. In particular, let $y_{k,m}(t)=1$ if UAV $k$ is scheduled to use sub-channel $m$ at time $t$, and $y_{k,m}(t)=0$ otherwise. Given that the spectrum holes are allocated to the requesting SUs based on the sub-channel availability, we incorporate the sensing and access costs to maximize the overall EE of the system. Therefore, we have:
\begin{equation}
\label{eqn:op3} 
\begin{cases}
\mathop{\mathrm{max}}\limits_{\{y_ {k,m}(t)\}} & \mathbb{E} \big\{ \sum\limits_{t,k,m} \frac{\ y_ {k,m}(t) \   c_ {k,m}(t) \ U_ {k,m}(t)} {y_ {k,m}(t) AC_ {k,m}(t) + SC_{k,m}(t) }  \big\}  \\
\text{subject to:} &  \sum_{m} \ y_ {k,m}(t) \ \leq 1 , \ \forall \ k=1,2,3, \dots K , \\
&  \sum_{k}  \ y_ {k,m}(t) \ \leq 1 , \ \forall \ m=1,2,3, \dots M , \\
&  \sum_{k,m} \ y_ {k,m}(t) \ \leq \ M -|\bm{z}(t)| , \ \\
&   y_ {k,m} (t) \in \{0,1\},  
\end{cases}
\vspace{-0.5mm}
\end{equation}
where $U_{k,m}(t)$, $SC_{k,m}(t)$, and $AC_{k,m}(t)$ are, respectively, the amount of data transmitted, the sensing cost, and transmission cost by the SU~$k$ on sub-band $m$. The constraints guarantee that each UAV is scheduled to use at most one sub-channel, while the total number of scheduled UAVs is at most equal to the number of detected spectrum holes at time $t$, which is $M-|\bm{z}(t)|$. In this paper, we use DNN at each UAV to detect spectrum holes that are fused to obtain $\bm{z}(t)$. To train the DNN models,  next we present an FL-based approach for distributed training of spectrum sensing models.

\section{Proposed FL-based Model Training for Spectrum Sensing}
\label{sec:proposedflsoln}
\subsection{Dataset and DNN Models}
\label{sec:sigmodel}
We assume that each UAV receives signals from more than one BS due to the fact that they operate at higher altitudes, which increases the chances of signal reception from multiple BSs.  
Furthermore, we assume that 
the cell bandwidth ${W}$ is partitioned into $M$ orthogonal sub-channels.
Then the total transmitted signal from a BS $\emph{b}$ across $M$ orthogonal sub-channels at any time~$t$ can be represented by the superposition principle as follows: 
\begin{equation}
    \label{eqn:txsig}
    {\bm{s}}_{b}(t) = \sum_{m=1}^{M}I_{b,m}(t) ~ \bm{v}_{b,m}(t), ~ ~ ~ \forall b \in \mathcal{B}, 
\end{equation}
where $I_{b,m}(t)=1$ if the $m$-th sub-channel of BS $b$ is occupied at time $t$, and $0$ otherwise. Moreover, $\bm{v}_{b,m}(t)$ represents the waveform on the $m$-th sub-channel.
As a result,  ${\bm{s}}_{b}(t)$ is the transmitted baseband waveform in digital domain.
Each UAV $k$ then receives a wideband signal from multiple BSs in a multi-path propagation environment, which can be expressed as follows:
\begin{equation}
    \label{eqn:sigrx2}
    \bm{r}_{k}(t) =\sum_{b=1}^{{B}} \bm{g}_{k,b}(t) * {\bm{s}}_{b}(t) + \bm{\eta}_{k}(t), ~ ~ ~ \forall k \in \mathcal{K}, 
\end{equation}
where $\bm{g}_{k,b}(t)$ represents the multi-path channel between BS~$b$ and UAV~$k$ and $\bm{\eta}_k (t)$ denotes the noise signal observed at UAV $k$. 
Therefore, the signal-to-noise ratio observed at UAV $k$ can be written as follows:  
\begin{equation}
    \label{eqn:sinreq}
    \text{SNR}_k(t) = \frac{||\sum_{b=1}^{{B}}  \bm{g}_{k,b}(t) * {\bm{s}}_{b}(t)|| ^2}{ \sigma_{k}^2(t)}, ~ ~ ~ \forall  k \in \mathcal{K},  
\end{equation}
where $\sigma_k ^2(t)$ represents the noise variance observed at UAV $k$ at time $t$. We use $P_k(t)$ to denote the total power received in UAV $k$ at time $t$, which is directly proportional to the signal generated as defined in~Eq.~\eqref{eqn:txsig}. We will use $P_k(t)$ in proportional weight scaling for FL training. 

To train the DNN models for predicting spectrum holes using raw I/Q samples, it has been shown that the characteristics of the wireless signal can be captured by observing only a portion of the signal waveform~\cite{uvaydov2021deepsense, chintareddy2023collaborative}.
Hence, from the received baseband signal ${\bm{r}}_k(t)$, we capture $J$ I/Q samples and store them locally. Therefore, the samples from baseband waveform collected at UAV $k$ are represented as $\bm{R}_k(t)$ given as follows:
\begin{equation}
    \label{eqn:sigrxk}
    {\bm{R}}_{k}(t) = \bm{\Tilde{R}}_{k}(t) +\bm{\Tilde{\eta}}_{k}(t), ~ ~ ~ \forall  k \in \mathcal{K}, 
\end{equation}
where $\bm{\Tilde{R}}_{k}(t)$ represents the~$J$ I/Q samples from the first term in Eq.~\eqref{eqn:sigrx2} and the second term represents~$J$ complex Gaussian noise samples.


In addition to the I/Q samples, we also need to store the true labels for channel occupancy at each UAV $k$ at time $t$. The channel occupancy vector ${\bm{h}}_k(t)$ is an $M$-dimensional vector, with each index indicating if a sub-channel $m$ is occupied or free at time $t$ and can be computed as follows:
\begin{equation}
h_{k,m}(t) =\begin{dcases}
1, &\sum_{b=1}^{B}~I_{b,m}(t) \geq 1;   \\
0, & \text{Otherwise.} 
\\
\end{dcases}  
\end{equation}
Note that ${\bm{h}}_k(t)$ observed at time $t$ would be the true label corresponding to the wideband received signal~$\bm{r}_k(t)$. The channel occupancy would remain unchanged for the stored $J$ I/Q samples~${\bm{R}}_{k}(t)$. We store (${\bm{R}}_{k}(t)$, ${\bm{h}}_k(t)$) as an input-output pair that will be used for the training of the FL model. For the sake of simplicity of notation, we represent the input-output pair as (${\bm{R}}_{k}$, ${\bm{h}}_k$). Note that for each $M$-dimensional channel occupancy vector $\bm{h}_k$, the input-output pair is treated as one data sample and the total I/Q dataset collected at UAV $k$ is denoted as follows: 
\begin{equation}
    \bm{D}_k = \bigl\{({\bm{R}_k^1}, \bm{h}_k^1), ({\bm{R}_k^2}, \bm{h}_k^2), \dots ({\bm{R}_k^{|\bm{D}_k|}}, \bm{h}_k^{|\bm{D}_k|})\bigr\}. 
\end{equation}
where $|\bm{D}_k|$ represents the total number of samples in the UAV $k$. These local datasets are used in FL-based training for spectrum hole detection.

In the FL setting, each UAV $k$ trains a local wideband spectrum sensing model whose parameters are denoted by $\bm{{\omega}_k}$. Hence, the primary objective of the local model is to find a mathematical function $f$($\bm{{\omega}}_k$, $\bm{R}_k$), that maps input I/Q samples $\bm{R}_k$ to $\bm{h}_k$, i.e.,
\begin{equation}
    f(\bm{{\omega}}_k, \bm{R}_k):\bm{R}_k \rightarrow \bm{h}_k. 
\end{equation}  
To this end, using the raw I/Q samples~($\bm{R}_k$) each UAV $k$ trains a local model that detects vacant sub-channels, 
such that the local loss function $L_k(\bm{{\omega}})$ minimizes the error between the true labels $\bm{h}_k$ and the predicted labels $\bm{\widehat{h}_k}$, as defined below:
\begin{equation}
\label{eqn:locloss}
L_{k}(\bm{\omega}) \triangleq \frac {1}{|\bm{D}_{k}|}\sum _{i=1}^{|\bm{D}_k|}l(f(\bm{\omega}_k, \bm{R}_k^i); \bm{{h}}_k^i)), 
\end{equation}
 where $l(.)$  is the loss function for computing the prediction loss in the supervised machine learning setting. Furthermore, $f(.)$ represents the predicted label for the sample ($\bm{R}_k^i$, $\bm{h}_k^i$) and $\bm{\omega}_k$ represents the local model parameters during training.



\begin{figure}[t]
\centering
    \includegraphics[width=0.5\textwidth]{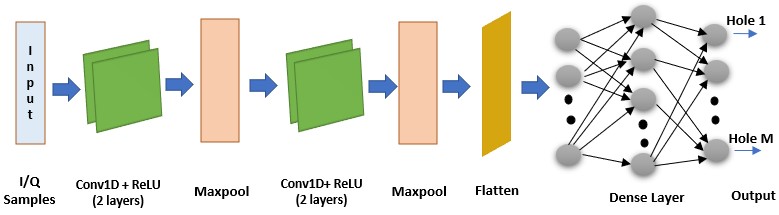}
    \caption{Multi-label classification using DNN.}
    \label{fig:cnn_with_dense}
    \vspace{-3mm}
\end{figure}

For each input sequence $\bm{R}_k^i$, we intend to obtain an $M$-dimensional binary vector $\bm{\widehat{h}_k}^i$ that represents the predicted spectrum holes. This is an instance of a multi-label classification problem for which we employ DNN. The DNN architecture considered is shown in Fig.~\ref{fig:cnn_with_dense}. The model accepts raw I/Q samples as input, which are then processed by two 1D convolutional (Conv1D) layers followed by a 1D maximum pooling layer (MaxPool1D). This layer pattern is repeated twice and one dense layer is followed by a sigmoid layer at the end. 

\subsection{Channel-Aware FL Aggregation Method}
Given the system model, we now introduce a framework for wideband spectrum sensing where multiple UAVs collaboratively participate in the FL. In such a distributed learning environment, we aim to learn a global statistical model at the central server. Given that each UAV $k$ trains a local model to identify the spectrum holes by minimizing the local loss function $L_k(\bm{\omega})$, in the context of FL, we would like to minimize the aggregated global loss function $L(\bm{\omega})$, as follows:
\begin{equation}  
\label{eqn:gloss}
\mathop {\min }_{\bm{\omega}} \Bigg\{L(\bm{\omega}) \triangleq \mathop  \sum _{k=1}^{{K}} \frac {|\bm{D}_{k}|}{{D}}~ L_{k}(\bm{\omega})\Bigg\}, 
\end{equation}
where ${D}$ = $\sum_{k=1}^{{K}}|\bm{D}_k|$ is the total size of data samples across the UAVs.

To solve the global loss function Eq.~\eqref{eqn:gloss}, the authors in~\cite{mcmahan2017communication} proposed FedAvg, an iterative aggregation algorithm where the global model aggregates the local model gradients and redistributes the global model weights to the local models. However, when the datasets of each UAV $k$ are of equal size, FedAvg assigns equal scaling factor of $\frac{1}{{K}}$ for all local gradients. However, in our considered multi-cell environment, the signal received at different UAV locations experiences different channel conditions, and the signal power received at different locations varies significantly. Hence, by assigning equal scaling weights for the local model gradients, the performance metrics at UAV locations with strong signal deteriorate. To compensate for this effect and improve performance at locations that receive better signal power, we propose a proportional weight scaling aggregation method for FL (pwFedAvg) that intuitively assigns smaller weights to UAVs with lower received signal power (i.e., poor channel conditions),  and larger weights to those UAVs with higher received signal power.

\begin{algorithm}[!t]
  \caption{Channel-Aware FL-Based Training}
  \label{alg:algo1}  
  \begin{algorithmic}[1]     
 \State Initialize the global model parameters $\bm{\omega}$ and local model $\bm{\omega_k}$,  $\forall k \in\mathcal{K}$; $T:$ Communication rounds.    
 \For{$t$ in $T$}
    \For{UAV $k$ in~$\mathcal{K}$}   
        \State Choose a batch of I/Q samples $\bm{\xi}_{k}^t$ $\subseteq \bm{D}_{k}$.
        \State Train the local model  for $\emph{E}$ epochs. 
   \EndFor \vspace{-1mm}     
    \State Send the gradients ${\nabla L_{k}(\bm{\omega}_k^{t};\bm{\xi}_{k}^t)}$ to the central server.
    \State Aggregate local gradients at the server per Eq.~\eqref{eqn:fl_agg}.
    \State Update the global model at the server per Eq.~\eqref{eqn:gm_update}.
    
    \State Update local models using the global model, i.e.,  ~ ~ ~$\bm{\omega}_k^{t+1}$ = $\bm{\omega}^{t+1}$.
    \EndFor   
\end{algorithmic}

\end{algorithm}

In particular, using the pwFedAvg, the central server aggregates the local gradients by assigning a weight proportional to their received signals as follows:
    \begin{equation}
    \label{eqn:fl_agg}
    {\nabla L(\bm{\omega}^t)} = \sum _{k=1}^{{K}} \frac{\alpha_k^t}{\alpha^t}~{\nabla L_{k}(\bm{\omega}_k^{t}; \bm{\xi}_{k}^{t})}, 
\end{equation}
where $\alpha_k^t$ = $\sqrt{\bar{P}_k^t}$ and $\alpha^t$ = $\sum_{k=1}^{{K}}{\sqrt{\bar{P}_k^t}}$. Here, ${\bar{P}_k^t}$ represents the average received signal power at UAV \emph{k} for the batch of samples $\bm{\xi}_{k}^t$. Note that during the FL training process at time $t$, $\bm{\omega}_k^t$ and $\bm{\omega}^t$ denote the local and global model weights, respectively. Upon computing the global model gradient based on Eq.~\eqref{eqn:fl_agg}, the global model weights are updated as follows:
   \begin{equation}
   \label{eqn:gm_update}
    \bm{\omega}^{t+1}= \bm{\omega}^{t} - \gamma^t ~{\nabla L(\bm{\omega}^t)}, 
\end{equation}   
where $\gamma^t$ is the learning rate of the global model. The updated global model weights are sent to the clients to update their local model weights. 
The overall process of FL-based model training using the pwFedAvg approach is outlined in Algorithm~\ref{alg:algo1}.

\subsection{Convergence Analysis}
In this section, we present the convergence analysis for the pwFedAvg algorithm. To this end,  we first introduce the following assumptions. 

\noindent 
\emph{Assumption 1:} The loss function $L_k(.)$ is $L$-smooth, i.e., for all \emph{$\bm{u}$} and \emph{$\bm{v}$},
\begin{equation}
L_k(\bm{u}) - L_k(\bm{v}) ~\leq~(\bm{u} - \bm{v})^{{T}}~ \nabla L_{k}(\bm{v}) + \frac{L}{2}~|| \bm{u} - \bm{v} ||_2^2.
\end{equation}

\noindent 
\emph{Assumption 2:} For each $k$, $L_k(.)$ is $\beta$-strongly convex, i.e., for all \emph{$\bm{u}$} and \emph{$\bm{v}$},
\begin{equation}
L_k(\bm{u}) - L_k(\bm{v}) ~\geq~(\bm{u} - \bm{v})^{{T}}~ \nabla L_{k}(\bm{v}) + \frac{\beta}{2}~|| \bm{u} - \bm{v}||_{2}^2.
\end{equation}

\noindent 
\emph{Assumption 3:} Let $\bm{\xi}_{k}$ be the data samples chosen from $\bm{D}_k$. The variance of the stochastic gradients for each UAV $k$ is bounded i.e,
\begin{equation}
\mathbb{E}\big[||\nabla L_{k}(\bm{\omega}_k;\xi_k) -  \nabla L_{k}(\bm{\omega}_k)||_{2}^2\big] ~\leq \rho_{k}^2 ~ ~\forall~k \in \mathcal{K}. 
\end{equation}

Next, similar to~\cite{li2019convergence,yan2022performance}, we define two virtual sequences  to denote the aggregated full gradient and stochastic gradient respectively, as follows:
\begin{equation}
    \bar{\bm{a}}^t = \sum _{k=1}^{{K}} \frac{\alpha_k^t}{\alpha^t}~{\nabla L_{k}(\bm{\omega}_k^{t})}; ~ ~\bm{a}^t = \sum _{k=1}^{{K}} \frac{\alpha_k^t}{\alpha^t}~{\nabla L_{k}(\bm{\omega}_k^{t}; \bm{\xi}_{k}^{t})}.
    \label{eqn:vsequences}
\end{equation}
We also assume that $\mathbb{E}[\bm{a}^t] = \bar{\bm{a}}^t$. Given these assumptions, we have the following lemmas.  

\noindent 
\textbf{Lemma 1:} Let $\bm{\omega^*}$ = [$\omega_1^*$, $\omega_2^*$, \dots , $\omega_d^*$] be the weights of optimal global model, and $\bm{\omega}_k^*$ = [$\omega_{k,1}^*$, $\omega_{k,2}^*$, \dots , $\omega_{k,d}^*$] be the weights of optimal local model of UAV $k$. Here, $d$ represents the dimensions of the model weights. Then for each UAV $k$, the upper bound of the gap between the optimal global and local models can be shown as,
\begin{equation}
L_{k}(\bm{\omega}^*) - L_{k}(\bm{\omega}_k^*) \leq \tau, \\
\end{equation}
\text{where} $\tau = \max\limits_{k}\{{\frac{Ld}{2}}(\max\limits_{i}\{ |\omega_i^* - \omega_{k,i}^*|\})^2\}$. 

\noindent 
\textbf{Lemma 2:} The aggregated gradient is upper bounded as follows:
\begin{equation}
\mathbb{E}\big(||\bm{a}^t - \bm{\bar{a}}^t ||_2^2 \big) \leq \sum_{k=1}^{{K}} \bigg(\frac{\alpha_k^t}{\alpha^t}\bigg)^2\rho_k^2. 
\end{equation}

\noindent 
 \textbf{Lemma 3:} Let the constants $\kappa$ and $\gamma^t$ satisfy $\frac{1}{\kappa}\leq \gamma^t$. Then, we can show that: 
\begin{equation}
\mathbb{E}\big[||\bm{\omega}^{t+1} - \bm{\omega}^* ||_2^2 \big] \leq (1-\beta \gamma^t) ||\bm{\omega}^{t} - \bm{\omega}^* ||_2^2 + (\gamma^t)^2 G^t,
\end{equation}
where $G^t = 2\kappa\tau + \sum_{k=1}^{{K}} \bigg(\frac{\alpha_k^t}{\alpha^t}\bigg)^2 \big[\rho_k^2-2L\tau \big]$.

\noindent 
\textbf{Theorem 1:} Given that $\kappa\leq \gamma^t = \frac{1}{\beta t + L}$, the optimality gap for the proposed pwFedAvg satisfies the following:
\begin{equation}
    \begin{split}
    \mathbb{E}[L(\bm{\omega})^T] - L^* \leq \frac{L}{\beta T + 2L} \bigg[ \frac{2G}{\beta} + L ||\bm{\omega}^0 - \bm{\omega}^* ||_2^2\bigg]     
    \end{split}, 
\end{equation}
where $G= \max_{t}\{G^t\}$ and $G^t$ is as defined in {Lemma $3$}. 
Therefore, we show that the convergence of our proposed method is $\mathcal{O}(\frac{1}{T})$. All of the proofs are presented in Section~\ref{sec:appendix}.

\section{Dynamic Spectrum Scheduling Using RL}
\label{sec:DSARL}
\begin{figure}[t!]
\includegraphics[width=\linewidth]{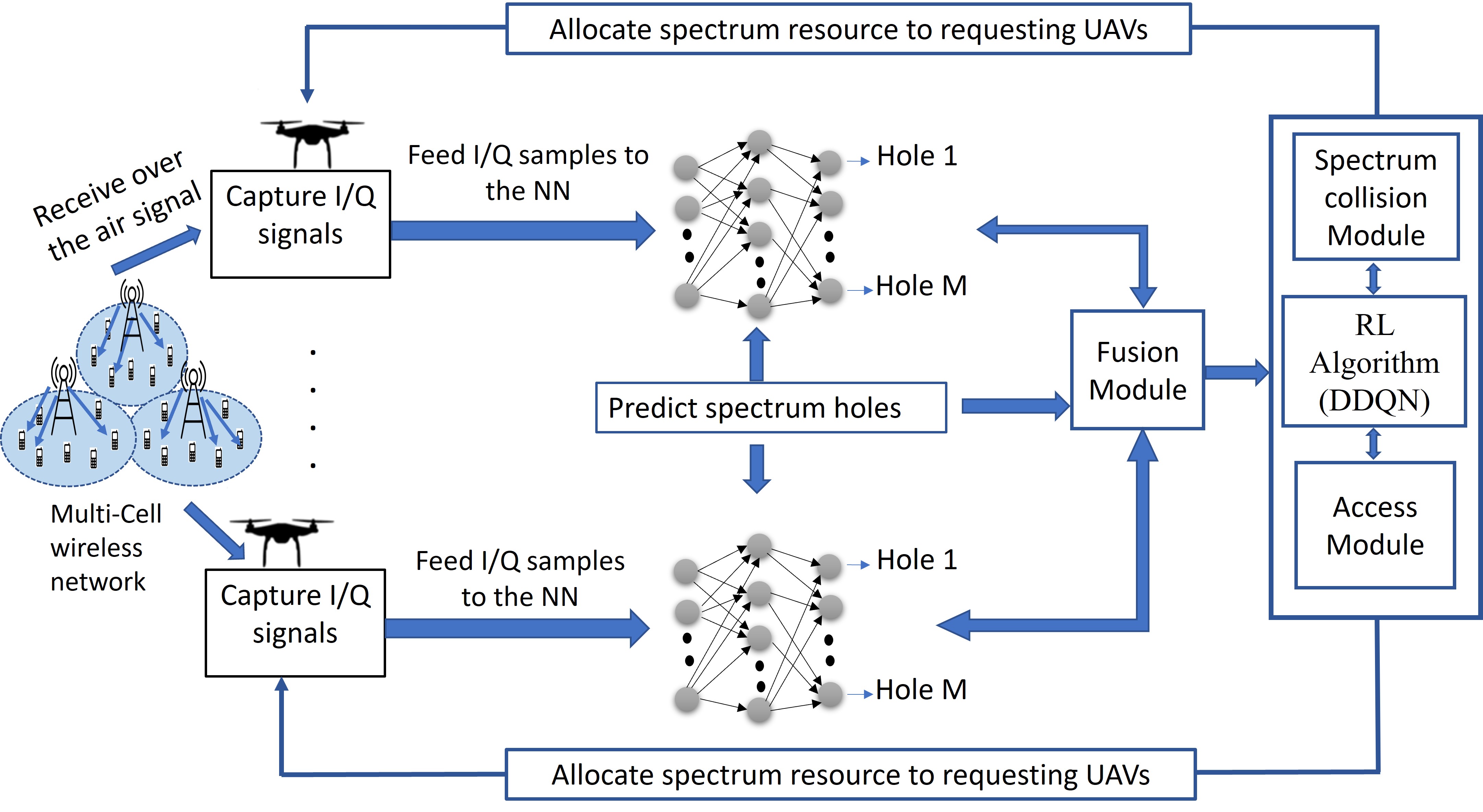}
  \vspace{-.5mm}
  \caption{Joint spectrum inference and spectrum scheduling.}
  \label{fig:testphase}
\end{figure}

Once the DNN models are trained using our proposed pwFedAvg, they output their spectrum hole predictions, which are then fused at the fusion module, as described in Section \ref{sec:sysmodel}. The identified spectrum holes will be allocated to requesting UAVs. The integrated system model of collaborative spectrum sensing and scheduling is shown in Fig.~\ref{fig:testphase}, with the overall algorithm described in Algorithm~\ref{alg:algo2}. 
For spectrum scheduling, we note that
the optimization problem in Eq.~\eqref{eqn:op3} is a fractional integer programming problem, which is NP-hard in general. If we consider maximizing the numerator alone, which is the total utility $U(t)$ of the UAVs over all sub-channels, the problem will become an integer programming problem. In this case, the utility would depend on the spectrum usage pattern by the PUs, which is captured by $c_{k,m}(t)$ as well as the channel condition between the BSs and UAVs that determine the amounts of transmitted data $U_{k,m}(t)$. To tackle this utility optimization problem,  we model the channel occupancy $\bar{z}_m(t)$ as a Markov process, enabling us to use an MDP formulation to solve this problem~\cite{sutton2018reinforcement} and develop a dynamic spectrum scheduling for the SUs.  
\label{sec:DSA}
As we assume that there exist $M$ sub-channels in the system, each sub-channel can be modeled as an independent two-state Markov chain. The transition probability function $\bm{P}$ can then be viewed as a set of transition probability matrices \{$\bm{P}_m$\} for each sub-channel that captures the randomness in the assumed multi-user multi-channel environment. Therefore, we formulate the total utility of the SUs as a traditional MDP governed by the tuple ($\mathcal{S}$, $\mathcal{A}$, \{$\bm{P}_m$\}, U, $\gamma$), consisting of the set of states $\mathcal{S}$, set of actions $\mathcal{A}$, a transition probability function \{$\bm{P}_m$\}, a reward function $U$, and a discount factor $\gamma$. To solve an MDP using RL, an agent learns to make decisions in an uncertain environment by maximizing a cumulative reward over a sequence of actions. Specifically, the agent interacts with an environment by taking actions that transition the system from one state to another, and the agent receives a reward that is commensurate with the merit of the action. The discount factor determines the relative importance of immediate and future rewards.  

\newcounter{phase}[algorithm]
\newlength{\phaserulewidth}
\newcommand{\setphaserulewidth}{\setlength{\phaserulewidth}}
\newcommand{\phase}[1]{%
  \Statex\leavevmode\llap{\rule{\dimexpr\labelwidth+\labelsep}{\phaserulewidth}}\rule{\linewidth}{\phaserulewidth}
  \Statex\strut\refstepcounter{phase}\textit{Phase~\thephase~--~#1}
  \vspace{-0.5ex}\Statex\leavevmode\llap{\rule{\dimexpr\labelwidth+\labelsep}{\phaserulewidth}}\rule{\linewidth}{\phaserulewidth}}
\makeatother

\setphaserulewidth{.3pt}

\begin{algorithm}[t!]
  \caption{Collaborative Spectrum Sensing and Scheduling}
  \label{alg:algo2}  
  \begin{algorithmic}[1]     
    \phase{Spectrum Sensing and Broadcasting }
    \For{each UAV in~$\mathcal{K}$}
        \State Capture I/Q samples from over the air signal.
        \State Feed I/Q samples to the pre-trained ML model that predicts the spectrum holes $\bm{\widehat{h}}$. 
        \State Broadcast the individual spectrum hole observations  $\bm{\widehat{h}}(t)$ $\in \{0,1\} ^{1 \times M}$ to the central server.      
        \EndFor \vspace{-1mm}
    \phase{Spectrum Fusion and Scheduling}
     \State Apply fusion rule in Eq.~\eqref{eqn:fusion}  to predict spectrum holes $\bm{z}(t)$.
    \State Allocate a single spectrum hole to each requesting UAV using pre-trained RL algorithm, $y_{k,m}(t)$, such that the constraints in Eq.~\eqref{eqn:op3} are satisfied.
    \State UAVs are scheduled to transmit on the sub-channel allocated in the previous allocated time slot.
    
    \State Given the spectrum allocation $y_{k,m}(t)$ and spectrum access collision indicator $c_{k,m}(t)$, the total utility $U(t)$ can be computed using Eq.~\eqref{eqn:op3}.
\end{algorithmic}
\end{algorithm}




\textbf{DDQN-Based Spectrum Allocation.} 
One of the most popular RL methods is Q-learning~\cite{sutton2018reinforcement}.
The classical Q-learning is table-based, i.e. the values of the Q-function are stored in a table of size $|\mathcal{S}|$×$|\mathcal{A}|$. However, when the size of the state and action spaces is large, the complexity of tabular Q-learning becomes cumbersome. For example, with $M = 16$ sub-channels, the Q-table will be of size $65,537 \times 17$.
To address the complexity issue, we adapt the deep Q-learning approach in~\cite{van2016deep} to approximate the Q-function by a neural network $Q_{\bm{\theta}}$ called DDQN and train its weights $\bm{\theta}$ using experience replay. As the name suggests, we have two networks when using DDQN where, $Q_{\bm{\theta}}$ is called the primary network and $Q'_{\bm{\theta}}$ is called the target network and the weights of the target network are updated periodically. In the original DDQN, the weights of target network are directly copied from the primary network every few episodes. In DDQN-soft, the target networks are updated using polyak averaging to smoothly update the weights (``soft-update'')~\cite{van2016deep}.

\begin{figure}[!t]
 \centering
    \includegraphics[width=0.98\linewidth]
    {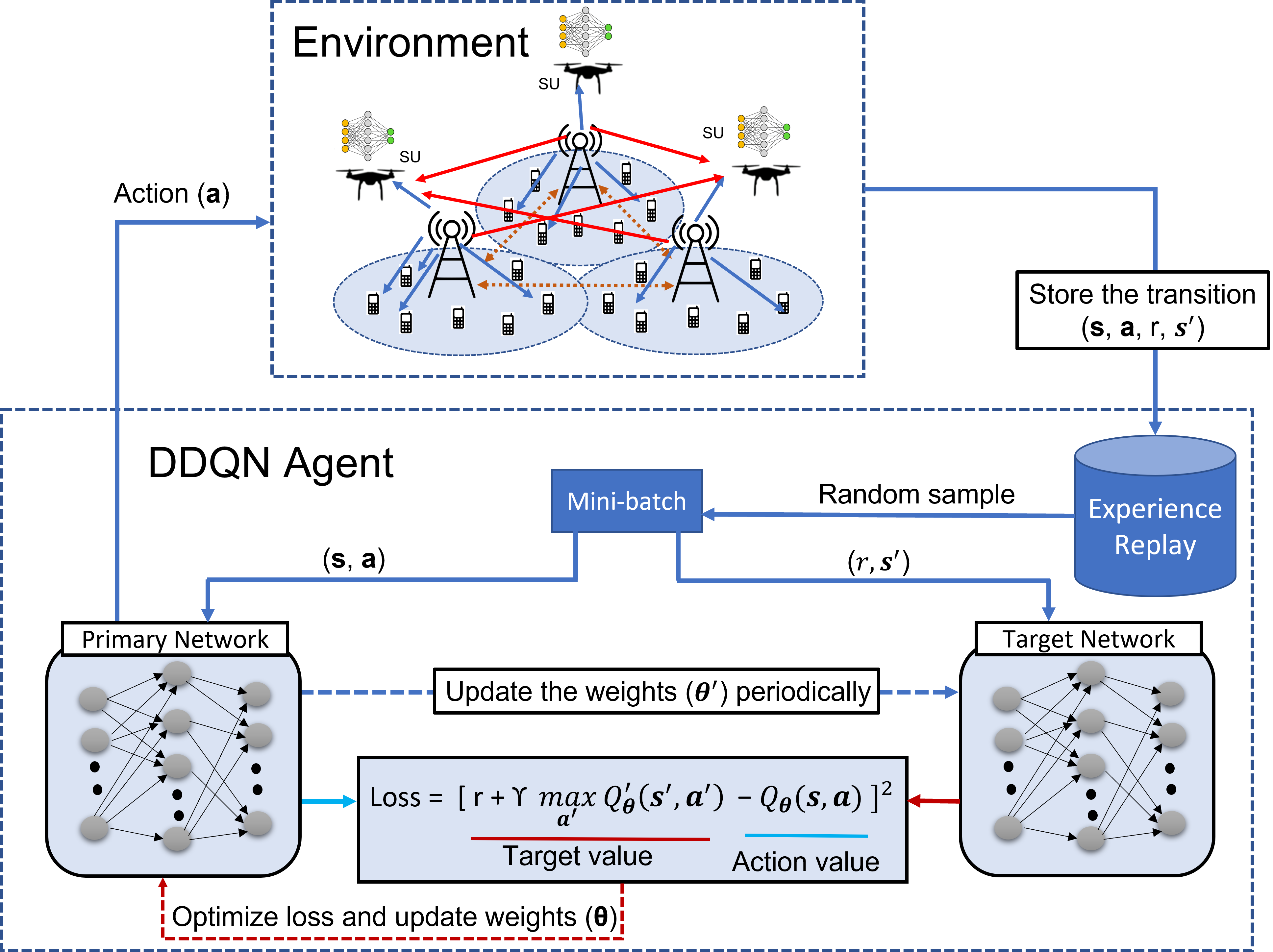}
     \vspace{-1mm}
    \caption{DDQN for spectrum allocation.}
    \label{fig:ddqndsa}  
    \vspace{-2mm}
\end{figure}

The input to the DDQN agent is a state \textbf{s} of size $1 \times M$. The output of the network is a vector of size $1 \times (M + 1)$ that contains the values of the Q-function with respect to state \textbf{s} and each of the $M + 1$ actions. In all hidden layers, we use the rectified linear unit~(ReLU) as an activation function. Given the neural networks input-output dimensions, the overall DDQN architecture and its interaction with the environment are shown in Fig.~\ref{fig:ddqndsa}. As shown, the major components are primary network, target network, experience replay, and the interaction with the environment to select an action.


To train the DDQN agent, the experiences are initially stored in the memory using $\epsilon$-greedy policy, that is, for a state $s_t$, an action $a_t$ is taken randomly with probability $\epsilon_t$, or taken greedily with probability $1$-$\epsilon_t$ from the current state of the DDQN network. Then, when we have sufficient samples in the memory a mini-batch of $\bm{X}$ experiences $\{(\textbf{s}_i,\textbf{a}_i,r_i,\textbf{s}_i ')\}_i \in \bm{X}_t$ is randomly sampled from the memory for every time step \emph{t} to train the neural networks. Here, $\bm{X}_t$ is the set of experiences currently available in the memory. Based on the selected mini-batch, we compute and update the weights $\bm{\theta}$ of the primary network $Q_{\bm{\theta}}$ that minimize the loss function $L_t({\bm{\theta}})$. Fig.~\ref{fig:ddqndsa} captures the overall DDQN architecture and the interaction of the agents with the environment~\cite{sutton2018reinforcement,van2016deep}. 

\section{I/Q DATASET GENERATION}
\label{sec:datasetgen}

Utilizing data-driven machine learning techniques for wide-band spectrum sensing requires substantial amounts of spectrum data. While obtaining raw I/Q signals over the air using physical hardware is the ideal scenario, the complexity of coordinating multiple UAVs in a specific environment for collaborative sensing poses significant challenges in achieving this objective.
Therefore, we resort to MATLAB's LTE toolbox to create the I/Q samples and employ ray-tracing methods to emulate the channel for generating synthetic datasets that closely mimic the data collection process through experimentation. The entire process of generating synthetic datasets is outlined below.

\begin{figure}[t!]
\centering  
\includegraphics[width=\linewidth]{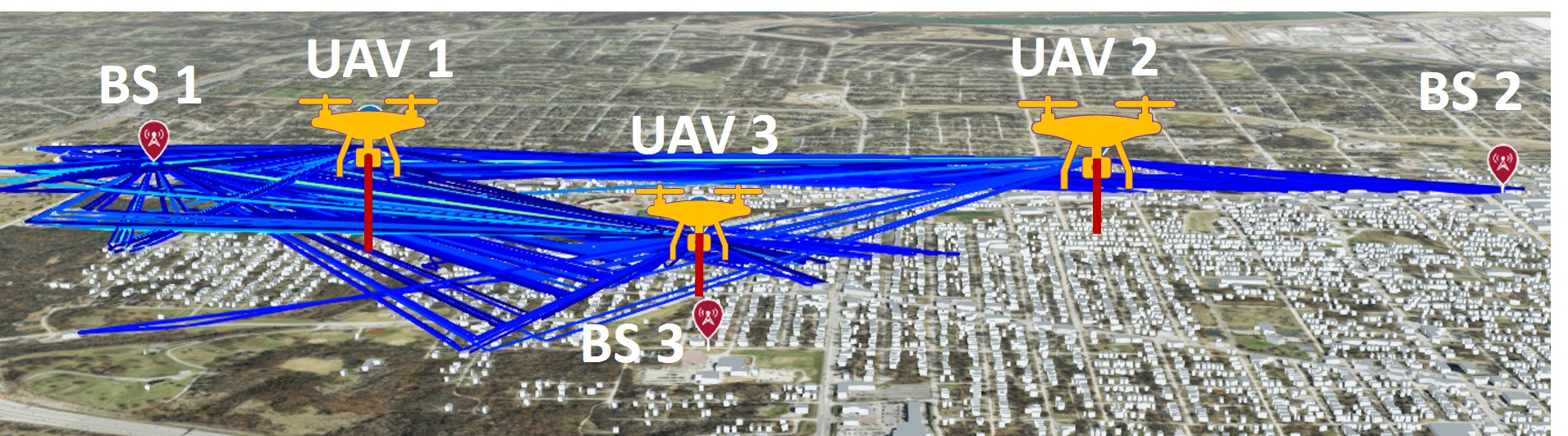}
    \caption{Ray-tracing simulation setup used for dataset generation. The plot illustrates the received signal paths at UAV location 1 from all three base-stations.}    
    \label{fig:simsetup}
\end{figure}

\textbf{Dataset Generation Methodology.}
As shown in Fig.~\ref{fig:simsetup}, we assume a multi-cell environment consisting of three neighboring cells with base-stations at the center of the cells. Without loss of generality, we simulate for one specific LTE band in the Kansas city area and obtain the location of the base-stations from cellmapper~\cite{cellmap}, an open crowd sourced cellular tower and coverage mapping service.  Furthermore, we assume there are three UAVs in the network operating at an altitude of $90$ meters. In this scenario, the base-stations act as the transmitter sites and the UAV locations as the receiver sites that collect the I/Q samples for wide-band spectrum sensing. 

Another important aspect in any wireless network is the wireless channel modeling. We use ray-tracing methods to incorporate the channel between the BS and the UAV. We incorporate both reflection and diffraction settings in ray-tracing to simulate a near real-world environment. This is in contrast to using channel models, which consider probabilistic channel models for line-of-sight (LoS) and non-LoS channel conditions. Since we use ray-tracing, we have the flexibility to incorporate different aspects of the environment like buildings and vegetation, permittivity and permeability of the materials, which further enhances the channel model. 

To mimic the real-world scenario for conducting ray-tracing experiments, we use OpenStreetMap, which is a free and open geographical database~\cite{OSM}. The evaluation area is a $3$~km $\times$ $3$~km area with buildings and vegetation. We utilize MATLAB's ray-tracer to emulate the wireless channel between considered UAVs and base-station locations. The ray-tracing simulation setup is outlined in Table~\ref{tab:rtsetup}. 
\begin{table}[!h]
    \centering
    \begin{tabular}{|c|c|}
    \hline
         \textbf{Parameter}& \textbf{Description}\\
         \hline
         Location & Kansas City \\
         \hline 
         Area & $3$~km x $3$~km \\
         \hline 
         Frequency& $1980$ MHz\\
         \hline
         Number of base-stations & $3$ \\
         \hline
         Number of UAVs & $3$\\
         \hline
         UAV Altitude & $90$~m \\
         \hline         
         Max. Number of Reflections & $5$\\ 
         \hline
         Max. Number of Diffractions & $2$\\
         \hline        
    \end{tabular}
    \vspace{2mm}
    \caption{Ray-tracing simulation setup.}
    \label{tab:rtsetup}
\end{table}

It is essential to highlight that this setup can be seamlessly adapted to accommodate varying numbers of LTE cells and UAVs as long as we are able to obtain the $3$D environment and load it into MATLAB. In our considered scenario, we assume the UAVs are stationary and are hovering in a fixed position. However, this simulation can be extended to incorporate the UAV flight trajectories by running additional ray-tracing experiments for each UAV way-point location in the UAV trajectory.

The MATLAB's ray-tracing toolbox effectively emulates the channel. Next, we utilize MATLAB's LTE Toolbox to generate the LTE waveform to extract the I/Q samples. For generating the LTE waveform, we assume that the entire cell bandwidth of $10$~MHz ($50$ resource blocks) is split into $16$ orthogonal sub-channels, each of size $3$ resource blocks. Typically, a base-station has the flexibility to assign either a single sub-channel or multiple sub-channels to a PU for transmitting user-specific data on the downlink shared channel. Additionally, various multiple access techniques can be employed to transmit data to different PUs in different time slots. However, during our dataset generation process, we do not consider primary user locations and how the base-station allocates user specific data to different PUs. At any given point in time, we take a snapshot of the entire cell bandwidth and identify which spectrum bands are occupied. Furthermore, when creating the downlink waveform, we omit the generation of UE specific reference signals to avoid mixing user-specific data with broadcast channels. Instead, we identify the appropriate indices and embed the LTE data samples into the downlink shared channel to generate the LTE waveform.

\begin{figure}[t!]
\centering  
\includegraphics[width=0.4\textwidth]{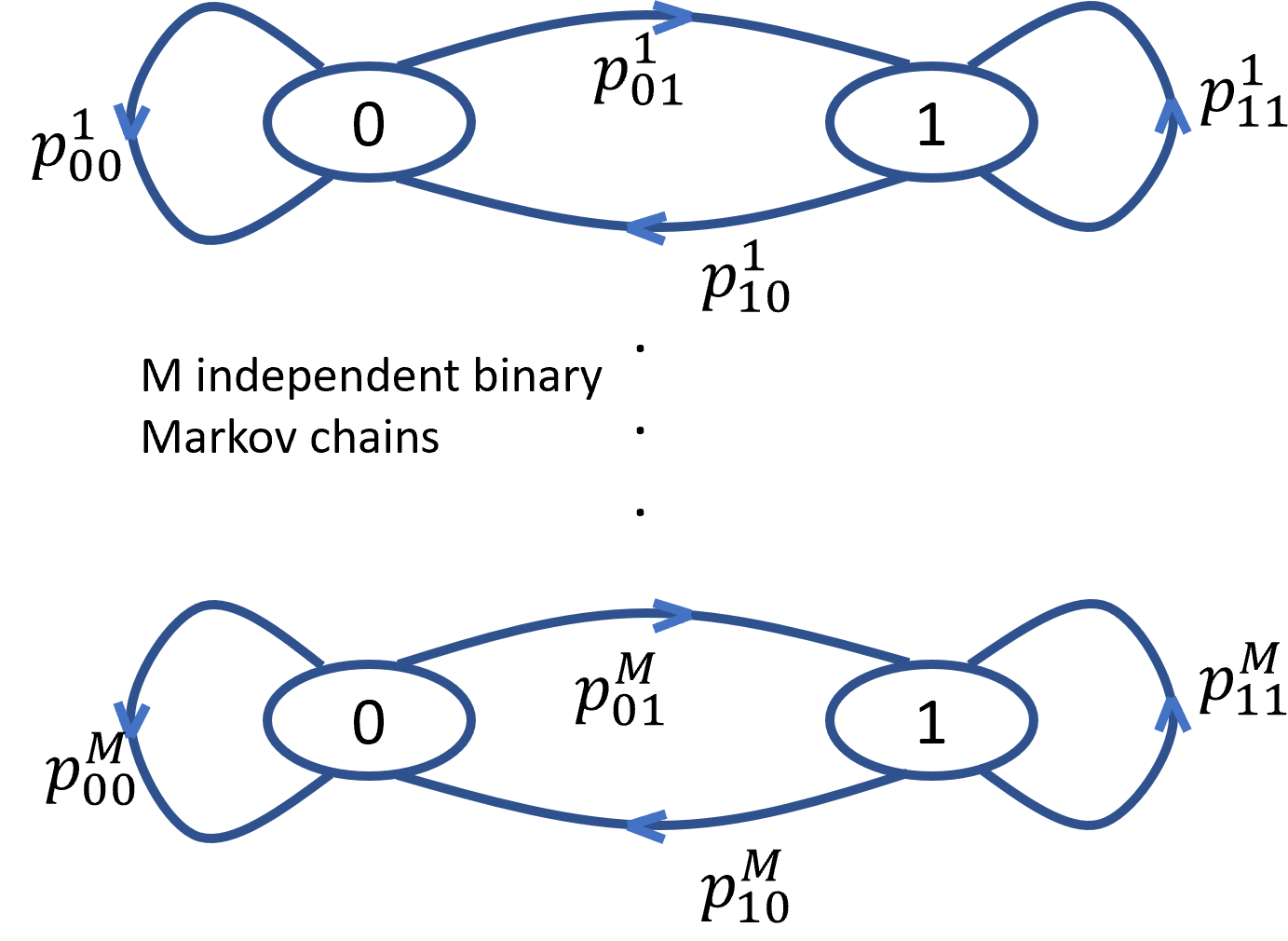}
    \vspace{-2mm}
    \caption{M independent Binary Markov chains. In our dataset generation, we set M = 16. }    
    \label{fig:markov}
\end{figure}

\begin{figure}[t!]
\centering  \includegraphics[width=0.9\linewidth]{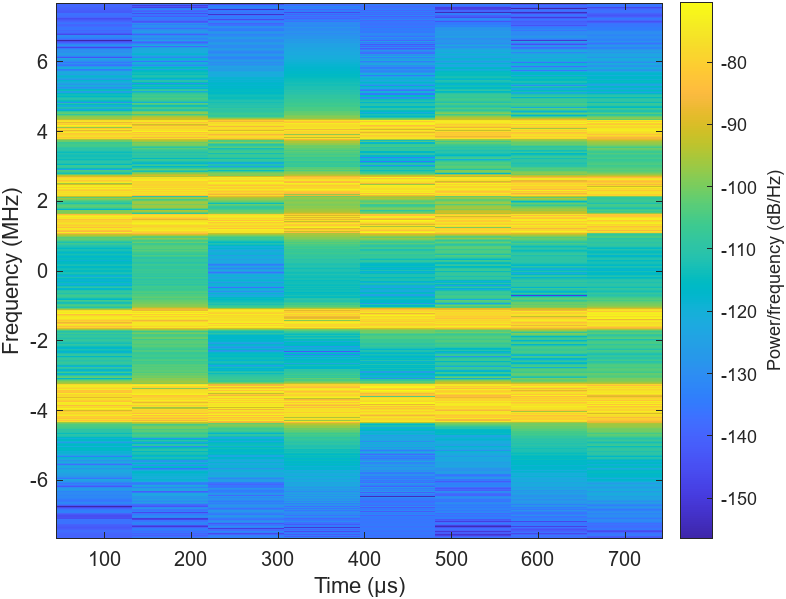}
  \caption{Spectrogram for a transmission on 6 sub-channels out of 16 sub channels.}
  \label{fig:s1clean1}
\end{figure}

\textbf{Modelling the channel occupancy.}
In our assumed scenario, each cell bandwidth is divided into $16$ sub-channels such that a binary flag $1$ indicates the sub-channel is allocated and $0$ represents the sub-channel is not allocated. Hence, each $16$-bit binary combination serves as a distinct true label for the channel occupancy. As a result, the base-station has the capability to generate $2^{16}$ unique labels, spanning from no sub-channel allocation to a fully busy cell site. For instance, in Fig.~\ref{fig:s1clean1} we show the spectrogram of one channel realization, where $6$ sub-channels are occupied out of $16$ sub-channels. Furthermore, we model the temporal dynamics of each sub-channel using a binary Markov chain,
as shown in Fig.~\ref{fig:markov}.
Thus, the channel occupancy for each sub-channel $m$ evolves according to a transition probability matrix $\bm{P}_m$. In this paper, we consider different transition probabilities for each sub-channel. Thus, the overall transition probabilities across $M$ sub-channels are denoted as follows:  
\begin{equation}    
\bm{P}=   \bigg\{\begin{bmatrix} p_{00}^1 & p_{01}^1 \\ p_{10}^1 & p_{11}^1 \end{bmatrix}, \cdots \cdots,  \begin{bmatrix} p_{00}^M & p_{01}^M \\ p_{10}^M & p_{11}^M
\end{bmatrix}\bigg\}.   
\end{equation}


\begin{figure}[t!]
\centering  
\includegraphics[width=0.9\linewidth]{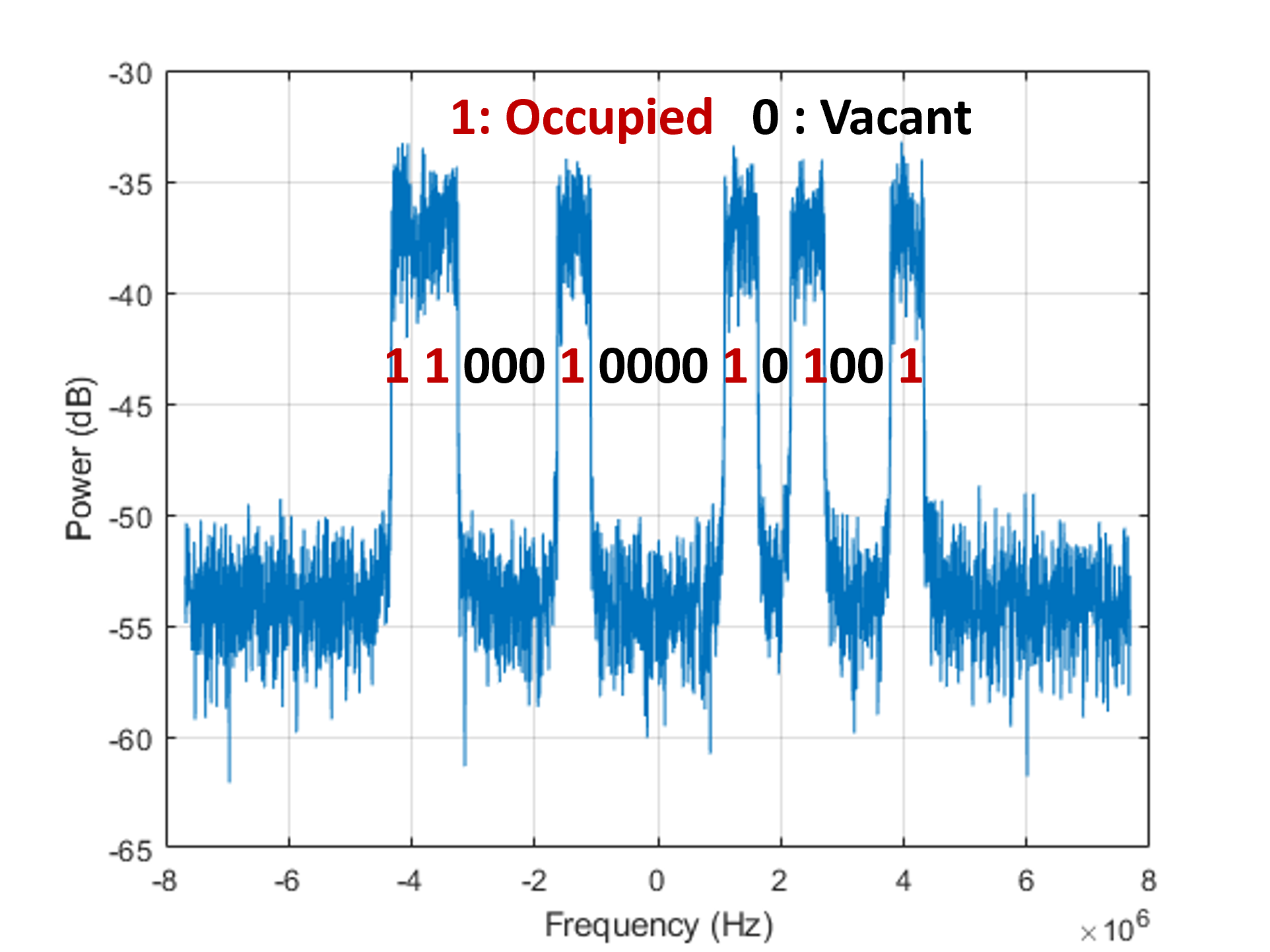}
  \caption{Power spectrum observed at UAV 1 for transmission on 6 sub-channels out of 16 sub channels, with channel impairments and for snr=-10dB; Vacant sub-channels are indicated as 0 and occupied sub-channels by 1.}
  \label{fig:s1noisy1}
\end{figure}

Further, we assume that all SUs are capable of receiving the waveform from all the base-stations, whose channel is modelled by the ray-tracing. In addition to the reflected paths received from the corresponding base-station in which the UAV is present, we also receive the waveform from the neighboring base-stations as shown in Fig.~\ref{fig:simsetup}. The received signal $\bm{r}_{k}(t)$ at each UAV $k$ can be written as superposition of wideband signals received from all base-stations as shown in Eq.~\eqref{eqn:sigrx2}. Furthermore, we vary the noise variance $\sigma_k ^2(t)$ at UAV $k$ such that the effective SNR varies from $-10~\text{dB}$ to $20~\text{dB}$ in steps of $10~\text{dB}$. For instance, in Fig.~\ref{fig:s1noisy1} we show the power spectrum of the received signal at UAV location~$1$. Ideally, we would like to capture the whole LTE frame corresponding to $10$~MHz LTE waveform. However, we only capture $32$ I/Q samples that provides a good trade-off between the computational complexity and the performance. In this context, for each SNR, we collect approximately~$6.8$ million I/Q samples. When considering all SNR levels and the UAV locations together, the total generated dataset is more than $80$ million I/Q samples, which will be publicly released along with all source codes. Generation of large-scale spectrum datasets for dynamic UAV environments enables us to evaluate the proposed data-driven collaborative wideband spectrum sensing and sharing, as described next. 


\section{Numerical Results}
\label{sec:results}



\begin{figure*}[t!]
\hspace{-1cm}
\centering
  \subfloat[\small{CL:~location~1.}]{
  \includegraphics[width=0.33\linewidth]{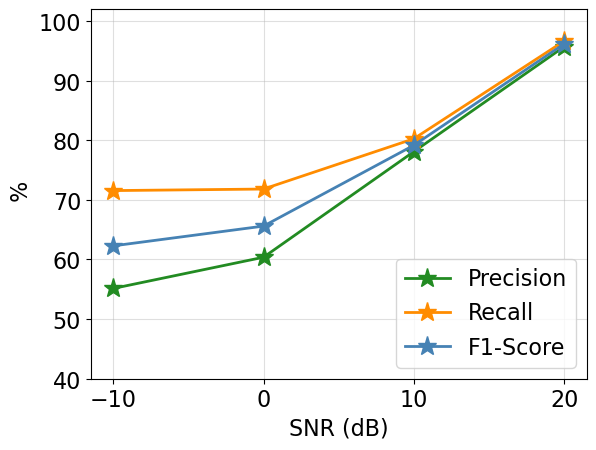}
  \label{fig:cluav1}
  }
 \subfloat[\small{CL:~location~2.}]{
 \includegraphics[width=0.33\linewidth]{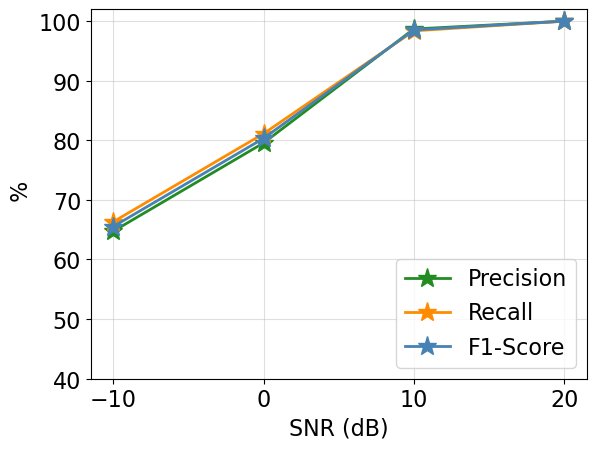}
  \label{fig:cluav2}
  }
\subfloat[\small{CL:~location~3.}]{
\includegraphics[width=0.33\linewidth]{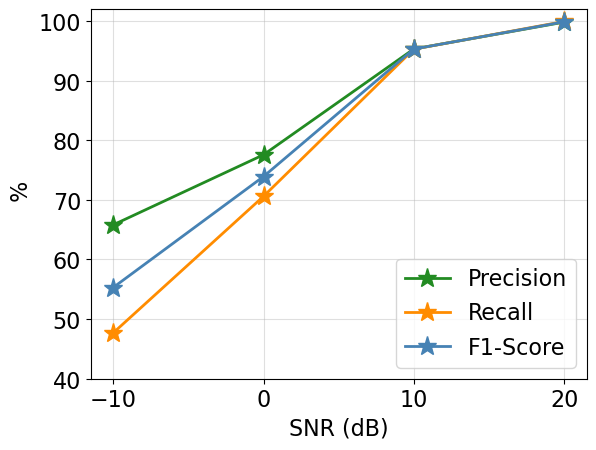}
  \label{fig:cluav3}
}
\caption[=]{\small{~\textbf{Performance metrics obtained at UAV locations 1, 2, 3 in CL. The performance metrics improve as the SNR observed by the UAV increases.}}}
\label{fig:cluav}
\end{figure*}

\begin{figure*}[!htb]
\hspace{-1cm}
\centering
\subfloat[\small{LL:~location~1.}]{  \includegraphics[width=0.33\linewidth]{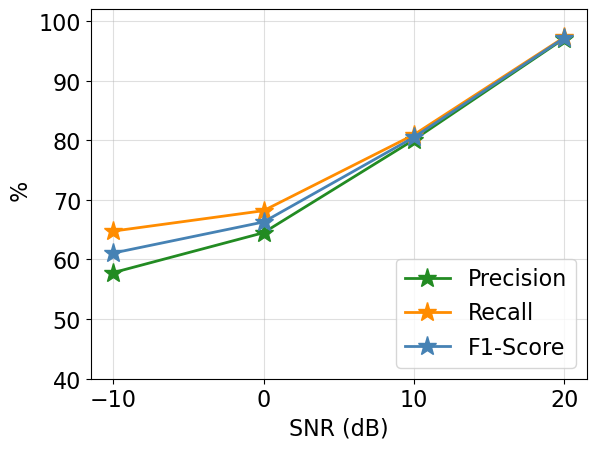}
  \label{fig:lluav11}
}
\centering  
\subfloat[\small{LL:~location~2.}]{
\includegraphics[width=0.33\linewidth]{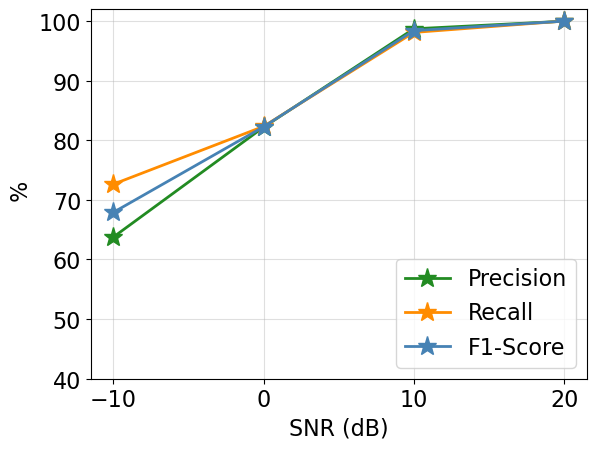}
  \label{fig:lluav22}
}
\centering 
\subfloat[\small{LL:~location~3.}]{
\includegraphics[width=0.33\linewidth]{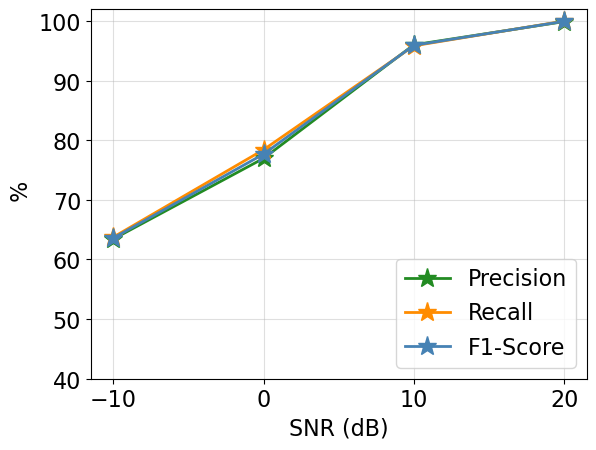}
\label{fig:lluav33}
}
\caption[=]{\small{~\textbf{Performance metrics obtained at UAV locations 1, 2 and 3 for testing their respective local models. The performance metrics improve as the SNR observed by the UAV increases.}}}
\label{fig:lluav}
\end{figure*}






 In this section, we first present our target performance metrics, followed by a discussion of the results on spectrum sensing for different ML configurations. Next, we present the results of collaborative spectrum inference and fusion followed by spectrum access using RL.

\textbf{Performance metrics.}
As mentioned earlier, detecting spectrum holes aligns with the framework of a classical multi-label classification problem, where each sub-channel represents a label. We utilize Precision, Recall, and F1-score as metrics to evaluate the classifier's performance for each sub-channel by constructing a confusion matrix. Although we can calculate these performance metrics for each sub-channel individually, it would be advantageous to have an average performance assessment across all $16$ sub-channels~\cite{grandini2020metrics}. In this paper, we consider the micro-averages for Precision, Recall and F1-score to concretely capture the sensing performance across the $16$ sub-channels as follows:
\begin{equation}
\text{Precision} = \frac{ \sum_{m=1}^{M} \text{TP(m)}}{\sum_{m=1}^{M} \text{TP(m)} + \text{FP(m)}},    
\end{equation}
\begin{equation}
\text{Recall}= \frac{ \sum_{m=1}^{M} \text{TP(m)}}{\sum_{m=1}^{M} \text{TP(m)} + \text{FN(m)}},    
\end{equation}
\begin{equation}
     \text{F1-score} = \frac{\text{2(Precision . Recall)}}{ \text{Precision} + \text{Recall} }.
\end{equation}
where TP, FN, FP accounts for the number of true positives, false negatives, and false positives, respectively.

\begin{figure*}[!ht]
\hspace{-1cm}
\centering
\subfloat[\small{LL model 1 tested at location 1.}]{  \includegraphics[width=0.33\linewidth]{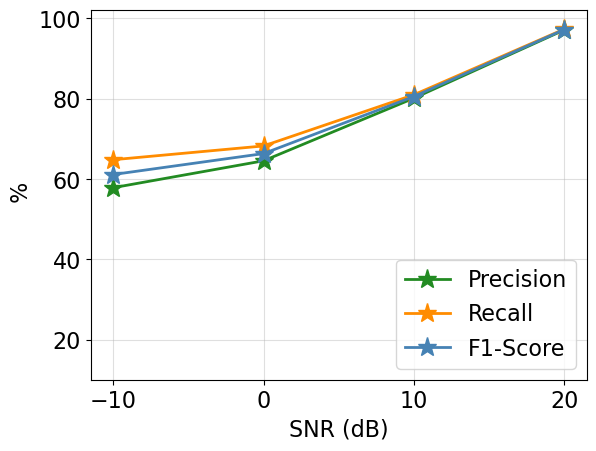}
  \label{fig:llgenuav11}
}
\centering  
\subfloat[\small{LL model 1 tested at location 2.}]{
\includegraphics[width=0.33\linewidth]{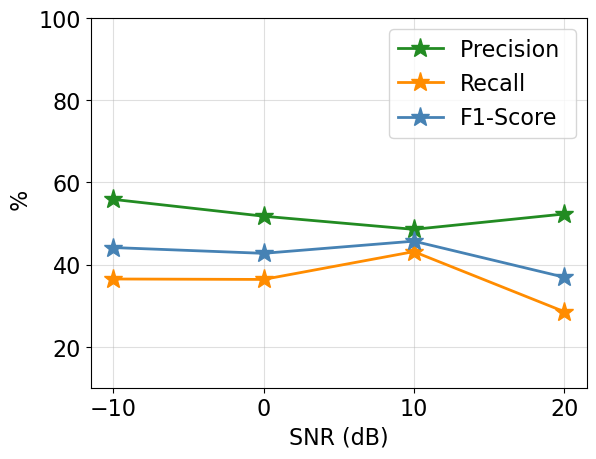}
  \label{fig:llgenuav12}
}
\centering 
\subfloat[\small{LL model 1 tested at location 3.}]{
\includegraphics[width=0.33\linewidth]{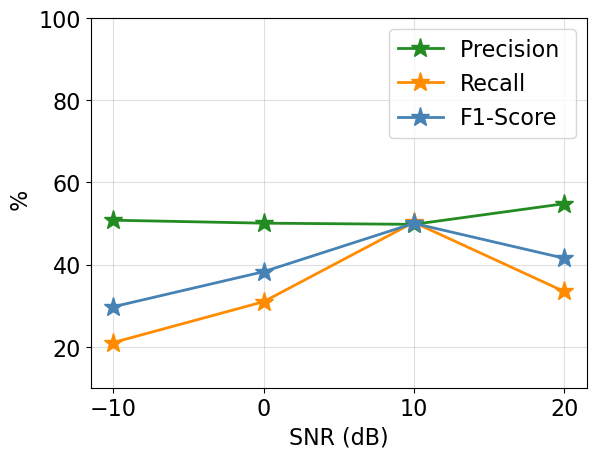}
\label{fig:llgenuav13}
}
\caption[=]{\small{~\textbf{Performance metrics obtained for testing UAV model 1 at UAV locations 1, 2 and 3. The local model trained at UAV 1 does not generalize for locations 2 and 3.}}}
\label{fig:llgenuav}
\end{figure*}
\begin{figure*}[t!]
\hspace{-1cm}
\centering
  \subfloat[\small{FL-FedAvg:~location~1.}]{
  \includegraphics[width=0.33\linewidth]{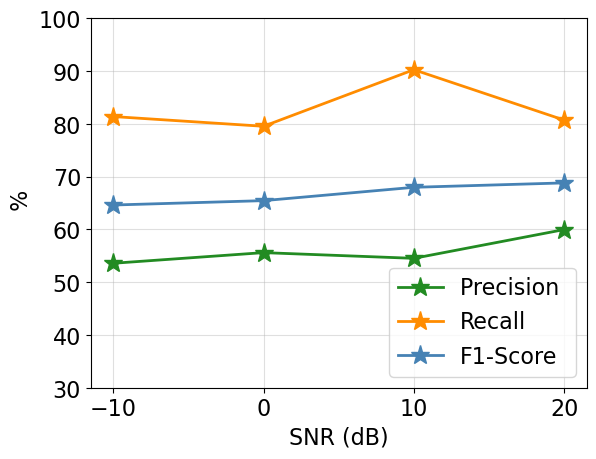}
  \label{fig:fluav1}
  }
 \subfloat[\small{FL-FedAvg:~location~2.}]{
 \includegraphics[width=0.33\linewidth]{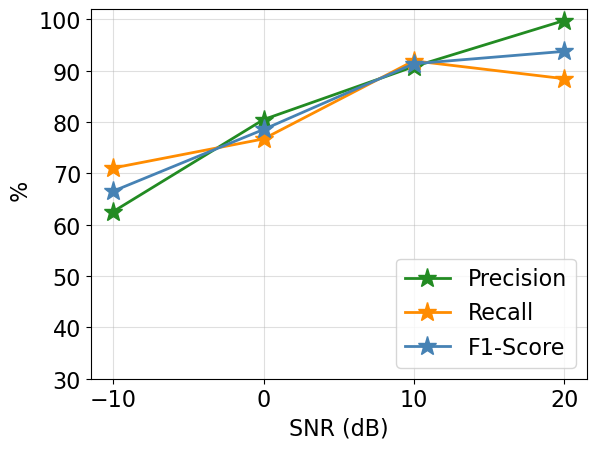}
  \label{fig:fluav2}
  }
\subfloat[\small{FL-FedAvg:~location~3.}]{
\includegraphics[width=0.33\linewidth]{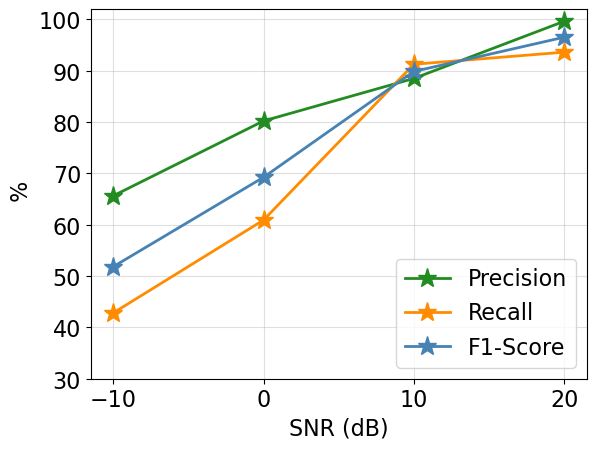}
  \label{fig:fluav3}}

\caption[=]{\small{~\textbf{Performance metrics obtained at UAV locations 1, 2 and 3 in FL using FedAvg. The performance metrics for locations 2 and 3 improve, but UAV location 1 remain almost the same.}}}
\label{fig:fluavfedavg}
\end{figure*}
\begin{figure*}[t!]
\hspace{-1cm}
\centering
  \subfloat[\small{FL-pwFedAvg: location 1.}]{
  \includegraphics[width=0.33\linewidth]{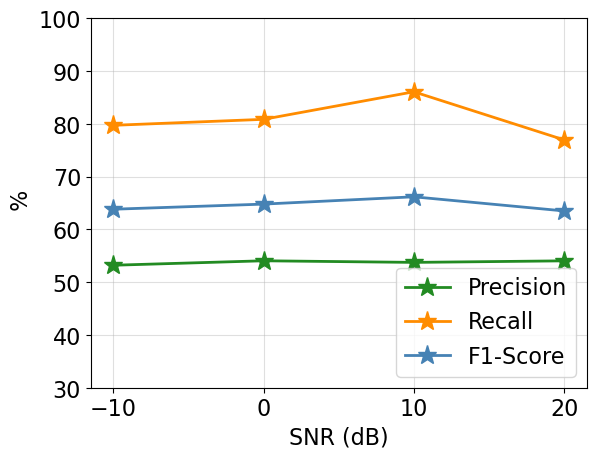}
  \label{fig:fluav11}
  }
 \subfloat[\small{FL-pwFedAvg: location 2.}]{
 \includegraphics[width=0.33\linewidth]{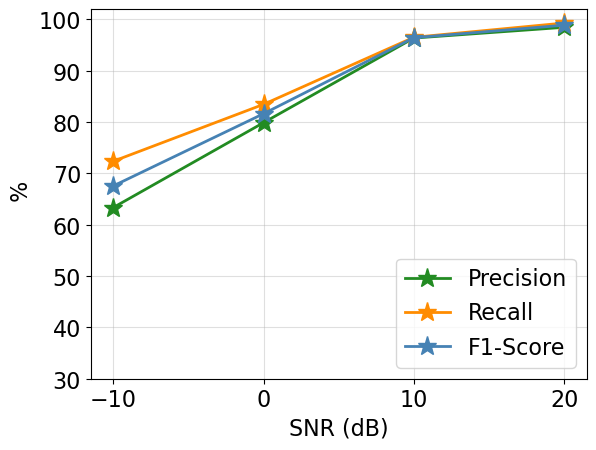}
  \label{fig:fluav22}
  }
\subfloat[\small{FL-pwFedAvg: location 3.}]{
\includegraphics[width=0.33\linewidth]{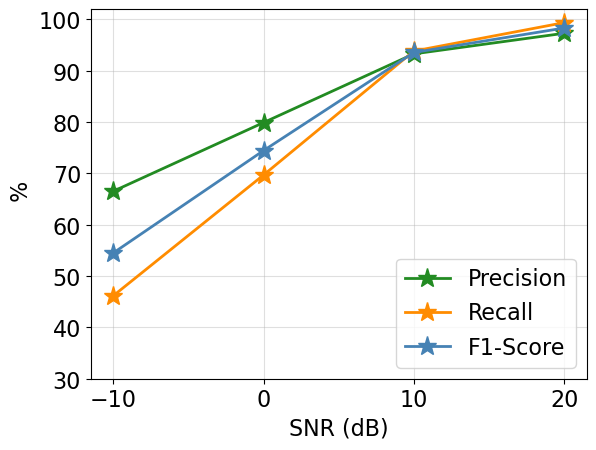}
  \label{fig:fluav33}}

\caption[=]{\small{~\textbf{Performance metrics obtained at UAV location 1, 2, 3 in FL using pwFedAvg. The performance metrics for locations 2 and 3 improve as snr increases and are higher than FedAvg.}}}
\label{fig:fluavpwfedavg}
\end{figure*}

\subsection{Model Training with Distributed UAVs}

As previously stated, we model wideband spectrum sensing, aiming to identify spectrum holes from the given I/Q samples as inputs to the ML model. In this context, we explore three configurations: centralized learning (CL),~local learning (LL),~and federated learning (FL), for training and testing the wideband spectrum sensing model. In each of these configurations, we use $70$\% of the dataset to train the model and $30$\% for spectrum inference purposes. Next, we present the results obtained for each configuration.

\textbf{Centralized Learning (CL)}  is a technique in which it is assumed that all the data collected at different locations are aggregated at one central server and are readily available to train the ML model.
The trained CL model is loaded on the UAVs for testing purposes, and the performance metrics computed at different UAV locations are shown in Fig.~\ref{fig:cluav}. As shown in Fig.~\ref{fig:cluav2}, \ref{fig:cluav3}, the performance metrics computed at UAV locations~$2$ and $3$ improves as SNR increases and attains near optimal performance around SNR~$20$ dB. However, as it can be seen in Fig.~\ref{fig:cluav1}, at UAV location $1$, the metrics are saturated at~$96$\%. Since CL accumulates all the datasets, the distributions of datasets at different locations are incorporated into the CL model, and thus it generalizes well to different locations, as shown in Fig.~\ref{fig:cluav}. However, the caveat of CL is the need to aggregate all datasets in one location to train a single model. On the other hand, we can consider local learning.

\begin{figure*}[t!]
\hspace{-1cm}
\centering
  \subfloat[\small{Comparison of F1-Score at location 1.}]{
  \includegraphics[width=0.33\linewidth]{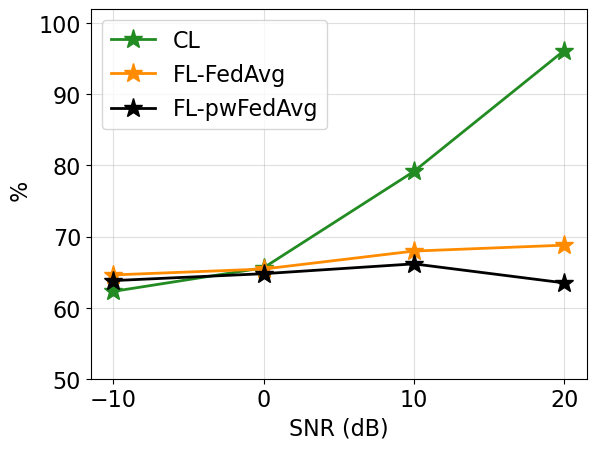}
  \label{fig:flllcluav1}
  }
 \subfloat[\small{Comparison of F1-Score at location 2.}]{
 \includegraphics[width=0.33\linewidth]{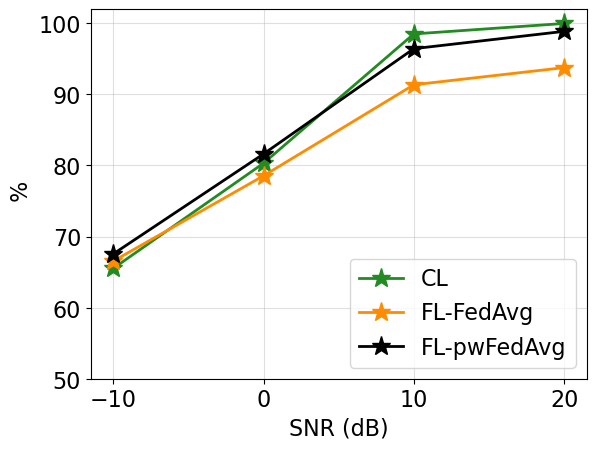}
  \label{fig:flllcluav2}
  }
\subfloat[\small{Comparison of F1-Score at location 3.}]{
\includegraphics[width=0.33\linewidth]{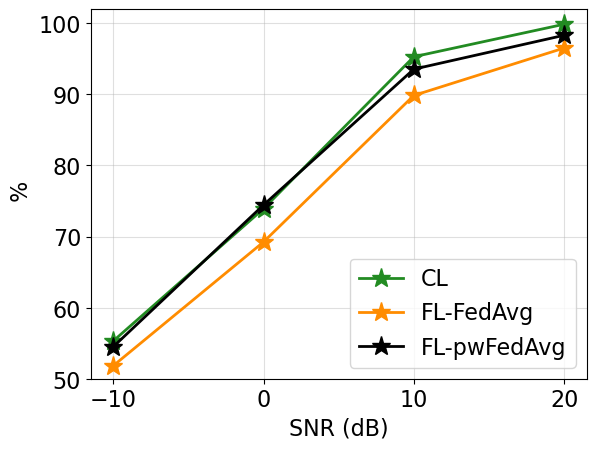}
  \label{fig:flllcluav3}}

\caption[=]{\small{~\textbf{Comparison of F1-score for CL, FL-FedAvg, FL-pwFedAvg. We improve the performance for locations 2 and 3 by proportionally scaling their weights as shown in Eq.~\eqref{eqn:fl_agg}}}. }
\label{fig:flllcluav}
\end{figure*}

\textbf{Local Learning (LL)} is a ML technique in which each UAV trains a model with its own local data, without sharing the dataset or model parameters with a central server or other UAVs. Hence, LL characterizes the performance of the model at a particular location. As shown in Fig.~\ref{fig:lluav}, the performance metrics improve as the SNR increases. Although LL tends to be a natural solution that provides insights into the performance of local models trained based on the local datasets, LL models at one particular location do not generalize to other locations.  
For example, in Fig.~\ref{fig:llgenuav}, when the local model trained at UAV location $1$ is tested at locations $2$ and $3$, the performance metrics are significantly lower than their individual performance metrics, as shown in Fig.~\ref{fig:lluav}.
This is one of the key observations that led us to explore federated learning that combines the advantages of both LL and CL to obtain a more generalized global model, without the need to accumulate all the datasets in one central location.

\textbf{Federated Learning} achieves trade-off between LL and CL, as it does not require aggregating the datasets in a central location; instead, the local model gradients are transferred to the central server for aggregation, and in return, the local models receive aggregated global weights as described in Algorithm~\ref{alg:algo1}. As such, the training process is similar to LL except that the local model weights are updated with the computed global weights iteratively, and by the end of the training process, all of the UAVs will have the same global model.
To investigate FL performance, we implement the FedAvg algorithm~\cite{mcmahan2017communication} and the results are presented in Fig.~\ref{fig:fluavfedavg}. From the results, we note that FedAvg achieves good performance only for the UAV locations~$2$ and $3$. Given the heterogeneous dataset collected at different UAV locations, the overall performance of FedAvg is limited by the UAV(s) that performs the worst. This is because FedAvg scales the weights of all local models equally. To reduce the impact of UAV locations with poor performance, our proposed pwFedAvg algorithm scales the weights of local models according to the received signal power. As shown in Fig.~\ref{fig:fluavpwfedavg} it is evident that our proportional weighting scheme improves the performance at locations $2$ and $3$. Furthermore, to have a fair comparison, we plot the F1-score for CL, FL-FedAvg and FL-pwFedAvg as shown in Fig.~\ref{fig:flllcluav}. With our proposed aggregating scheme (pwFedAvg), we improve the performance metrics at UAV locations $2$ and $3$, without significantly affecting location $1$ performance.

\begin{figure}[!t]
\hspace{-1cm}
\centering  
\includegraphics[width=.8\linewidth]{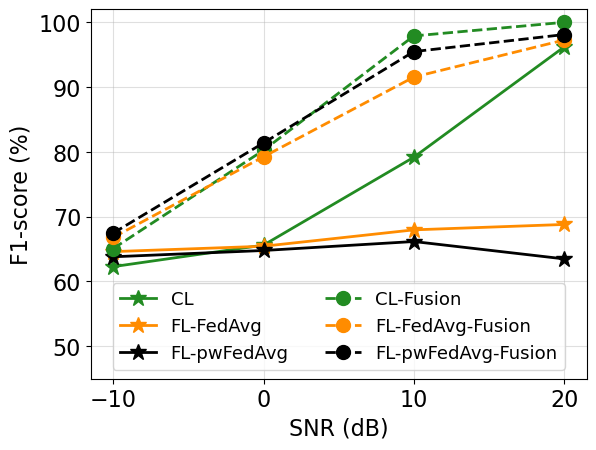}
    \caption{Comparison of F1-Score results with and without fusion for CL, FL-FedAvg and FL-pwFedAvg at UAV location 1. }    
    \label{fig:fusion2}
\end{figure}
\begin{figure}[t]
\hspace{-.8cm}
\centering
\subfloat[]{
 \includegraphics[width=0.5\linewidth]{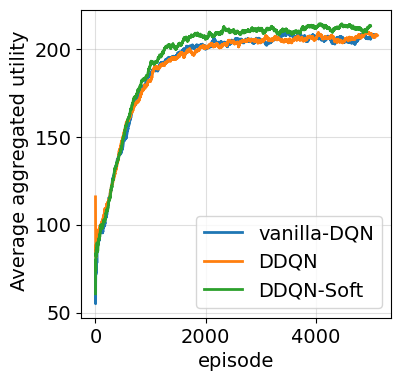}
    \label{fig:dqn-1UAV}
}
\subfloat[]{
\centering
  \includegraphics[width=0.5\linewidth]{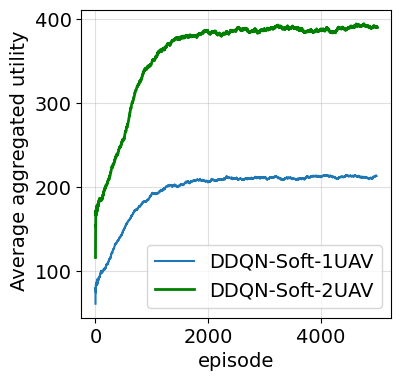}
    \label{fig:dqn-2UAV}
}
\caption{Training results for allocating spectrum holes. (a) to \textbf{one UAV}. (b) to \textbf{two UAVs}.}
\label{fig:RLtrainresult}
\end{figure}

\subsection{Collaborative Spectrum Inference results}
As shown in Fig.~\ref{fig:testphase}, we consider fusing the spectrum hole predictions from multiple UAVs.
This is motivated by the fact that individual sensing performance might fluctuate at different locations, which we observed in the CL, LL, FL settings. However, by applying fusion rules, we can significantly improve the overall performance, as shown by our results in Fig.~\ref{fig:fusion2}. 
From the results, we notice that the overall performance of all methods is significantly improved by fusion. Furthermore, the proposed pwFedAvg algorithm outperforms FedAvg, while achieving comparable results with respect to the CL method without the need to transfer all datasets to a central location. The comparison results for the spectrum fusion results at locations $2$ and $3$ are omitted for brevity, as they show similar trends.


\subsection{Spectrum Resource Allocation using RL}
As mentioned in Section~\ref{sec:DSARL}, we use deep Q-learning methods for allocating spectrum resources to the UAVs. 
In Fig.~\ref{fig:dqn-1UAV}, we compare the training performance of three variants of Q-learning methods for allocating a sub-channel to a single UAV whenever the fusion rule detects at least a single spectrum hole. It is observed that DDQN with soft update performs slightly better and converges earlier than DDQN and vanilla-DQN. 
Next, we extend the model to allocate spectrum holes to two UAVs. In this case, we have augmented the DDQN algorithm with soft update to generate two best actions.  From the results in Fig.~\ref{fig:dqn-2UAV}, we observe that the utility performance with two SUs is slightly less than two times of the performance with a single SU. 
We further note that this paper tries to explore the possibility of integrating spectrum sensing and sharing by making use of existing RL algorithms. Though we explored Q-learning techniques, different and other advanced RL algorithms can be integrated into the proposed framework.






\section{Conclusion}
\label{sec:conclusion}

In this paper, we developed a collaborative wideband spectrum sensing and sharing solution for networked UAVs. To train machine learning models for detecting spectrum holes, we explored the applications of FL and developed an architecture that integrates wireless dataset generation into the FL model training and aggregation steps. To this end, we proposed the pwFedAvg algorithm to incorporate wireless channel conditions and received signal powers into the FL aggregation algorithm. To further enhance the accuracy of the predicted spectrum holes by individual UAVs, we considered spectrum fusion at the central server. Additionally, by leveraging deep Q-learning methods, the detected spectrum holes are dynamically allocated to the requesting UAVs. To evaluate the proposed methods, we generated a near-realistic synthetic dataset using MATLAB LTE toolbox by incorporating base-station locations in a chosen area of interest, performing ray-tracing, and emulating the primary users channel usage in terms of I/Q samples. Based on the collected I/Q datasets, we investigated the performance of three model training algorithms, namely CL, LL, and FL. 
The numerical results demonstrated that the CL model generalizes well and performs better for all UAV locations, while the LL models showed poor generalization performance. 
Furthermore, the proposed pwFedAvg algorithm outperforms FedAvg while achieving comparable results with respect to the CL method without the need for sharing all datasets to a central location. From the fusion results, we noticed that the overall performance improved significantly for all learning configurations, and the implemented DDQN method can provide dynamic spectrum scheduling across requesting UAVs. In future work, we plan to expand the application of our developed solutions to other technologies and spectrum bands (beyond LTE), while incorporating realistic spectrum usage of the incumbent users in those bands (i.e., PUs).

\section{APPENDIX}
\label{sec:appendix}
\textbf{Proof of Lemma $1$.} 
 Since $\nabla L_k(\bm{\omega}_{k}^*)$ = $0$, assumption $1$ reduces to $L_{k}(\bm{\omega}^*)$ - $L_{k}(\bm{\omega}_k^*)$ $\leq$ $\frac{L}{2}$ $||\bm{\omega}^* - \bm{\omega}_k^* ||_2^2 $. 
Using the identities of vector norm and max-norm for a vector $\mathbf{x}$, $||\mathbf{x}||_2^2 \leq~d||\mathbf{x}||_\infty^2 = d (\max\limits_i |x_i|)^2$, we have: 
$$
\frac{L}{2} ||\bm{\omega}^* - \bm{\omega}_k^* ||_2^2 \leq \max\limits_{k}\{{\frac{Ld}{2}}(\max\limits_{i}\{ |\omega_i^* - \omega_{k,i}^*|\})^2\}, 
$$
which completes the proof. 

\noindent 
\textbf{Proof of Lemma $2$.}
 Using the definition of virtual sequences from Eq.~\eqref{eqn:vsequences}, we have: \\ 
$ \mathbb{E}\big(||\bm{a}^t - \bm{\bar{a}}^t ||_2^2 \big) $ 
\begin{equation}
\begin{split}
&\overset{\mathrm{(a)}}{=}\small{\mathbb{E}\big[\sum_{k=1}^{{K}} ||{\bigg(\frac{\alpha_k^t}{\alpha^t}\bigg) \big[\nabla L_{k}(\bm{\omega}_k^{t}; \bm{\xi}_{k}^{t})} - {\nabla L_{k}(\bm{\omega}_k^{t})\big] ||_2^2} \big] } \\
&\overset{\mathrm{(b)}}{\leq} \sum_{k=1}^{{K}} \bigg(\frac{\alpha_k^t}{\alpha^t}\bigg)^2 \mathbb{E}\big[||{\nabla L_{k}(\bm{\omega}_k^{t}; \bm{\xi}_{k}^{t})} - {\nabla L_{k}(\bm{\omega}_k^{t}) ||_2^2} \big] \\
&\overset{\mathrm{(c)}}{\leq} \sum_{k=1}^{{K}} \bigg(\frac{\alpha_k^t}{\alpha^t}\bigg)^2\rho_k^2,~ 
\end{split}
\end{equation}
where~(\emph{a})~is from~Eq.~\eqref{eqn:vsequences}, (\emph{b}) comes from Jensen's inequality, and (\emph{c}) is by applying Assumption $3$.\\
\\
\textbf{Proof of Lemma $3$.}
~\text{Using Eq.~\eqref{eqn:fl_agg} and Eq.~\eqref{eqn:gm_update}}, we have the following equation: \\
$ ||\bm{\omega}^{t+1} - \bm{\omega}^*||_2^2 = ||\bm{\omega}^{t} -\gamma^t \bm{a}^t - \bm{\omega}^* ||_2^2 $ 
\begin{equation}
\begin{split}
&=||\bm{\omega}^{t}-\gamma^t \bm{\bar{a}}^t +\gamma^t \bm{\bar{a}}^t-\gamma^t \bm{a}^t - \bm{\omega}^* ||_2^2 \\
&= \underbrace{||\bm{\omega}^{t}-\gamma^t \bm{\bar{a}}^t  - \bm{\omega}^* ||_2^2}_\text{$A_1$} + \underbrace{(\gamma^t)^2 ||\bm{ \bm{a}^t -\bar{a}}^t||_2^2}_\text{$A_2$} \\
&\underbrace{-2\gamma^t \langle\bm{\omega}^{t}-\gamma^t \bm{\bar{a}}^t  - \bm{\omega}^* ,\bm{a}^t -\bar{\bm{a}}^t \rangle}_\text{$A_3$}. \\
\end{split}
\end{equation}

\noindent 
Since $\mathbb{E}[\bm{a}^t] = \bar{\bm{a}}^t $, it can be seen that $\mathbb{E}[A3] =0$. \\
By expanding $\mathbb{E}[A_1]$, we have: 

\begin{equation}
\begin{split}
\mathbb{E}[A_1] &=\mathbb{E}\big[ ||\bm{\omega}^{t} - \bm{\omega}^{*} ||_2^2 + \underbrace{(\gamma^t)^2 ||\bar{\bm{a}}^t||_2^2}_\text{$A_{1, 1}$}\\ &-\underbrace{2~\gamma^t \langle\bm{\omega}^{t}- \bm{\omega}^* ,\bar{\bm{a}}^t \rangle}_\text{$A_{1, 2}$}\big].\\
\end{split}
\end{equation}

The bound on $A_{1, 1}$ term can be derived as follows:
$\mathbb{E}[A_{1, 1}]$
\begin{equation}
\begin{split}
&\overset{\mathrm{(a)}}{=}(\gamma^t)^2~\mathbb{E}\big[ ||\sum_{k=1}^{{K}} \frac{\alpha_k^t}{\alpha^t} \nabla L_{k}(\bm{\omega}_k^{t})||_2^2 \big]\\
&\overset{\mathrm{(b)}}{\leq} (\gamma^t)^2 \sum_{k=1}^{{K}} \bigg(\frac{\alpha_k^t}{\alpha^t}\bigg)^2 ||\nabla L_{k}(\bm{\omega}_k^{t})||_2^2\\
&\overset{\mathrm{(c)}}{\leq} 2L~(\gamma^t)^2 \sum_{k=1}^{{K}} \bigg(\frac{\alpha_k^t}{\alpha^t}\bigg)^2 \big( L_{k}(\bm{\omega}_k^{t}) - L_{k}(\bm{\omega}_k^{*}) \big), \\
\end{split}
\end{equation}
where~(\emph{a})~is from~Eq.~\eqref{eqn:vsequences}, (\emph{b}) comes from Jensen's inequality, and (\emph{c}) by applying Assumption~$1$ and $L$-smoothness property~\cite{boyd2004convex}. The bound for 
$A_{1, 2}$ term can be derived as follows: \\
$\mathbb{E}[A_{1, 2}]$ = $-2~\gamma^t ~\langle\bm{\omega}^{t}- \bm{\omega}^* ,\sum_{k=1}^{{K}} \frac{\alpha_k^t}{\alpha^t}~{\nabla L_{k}(\bm{\omega}_k^{t})} \rangle$

\begin{align*}
&=-2\gamma^t~\sum_{k=1}^{{K}} \frac{\alpha_k^t}{\alpha^t} ~\langle\bm{\omega}^{t}- \bm{\omega}^* ,~{\nabla L_{k}(\bm{\omega}_k^{t})} \rangle\\
&\overset{\mathrm{(a)}}{\leq} -2\gamma^t~\sum_{k=1}^{{K}} \frac{\alpha_k^t}{\alpha^t}\big[ L_{k}(\bm{\omega}_k^{t}) - L_{k}(\bm{\omega}^{*})  + \frac{\beta}{2}\bm{||\omega}^{t}- \bm{\omega}^*||_2^2\big] \\
&\overset{\mathrm{(b)}}{\leq} \underbrace{-2\gamma^t~\sum_{k=1}^{{K}} \frac{\alpha_k^t}{\alpha^t}\big( L_{k}(\bm{\omega}_k^{t}) - L_{k}(\bm{\omega}^{*}) \big)}_\text{$A_{1, 2, 1}$} - \beta\gamma^t~||\bm{\omega}^{t}- \bm{\omega}^*||_2^2,
\end{align*}
where~(\emph{a})~comes from Assumption $2$,~(\emph{b}) comes from the fact that $\sum _{k=1}^{{K}} \frac{\alpha_k^t}{\alpha^t}=1.$

Combining $\mathbb{E}[A_{1, 1}]$ and $\mathbb{E}[A_{1, 2, 1}]$, we have\\

$\mathbb{E}[A_{1, 1}]$ + $\mathbb{E}[A_{1, 2, 1}]$\\
\begin{equation}
\begin{split}
&\leq 2L~(\gamma^t)^2 \sum_{k=1}^{{K}} \bigg(\frac{\alpha_k^t}{\alpha^t}\bigg)^2 \bigg( L_{k}(\bm{\omega}_k^{t}) - L_{k}(\bm{\omega}^{*}) \\
&~ ~ ~ ~ ~ ~ ~ ~ ~ ~ ~ ~ ~ ~ ~ ~  ~ ~ ~ ~ ~ ~ ~ ~ ~ ~ ~ + L_{k}(\bm{\omega}^{*})- L_{k}(\bm{\omega}_k^{*}) \bigg)  \\
&-2\gamma^t~\sum_{k=1}^{{K}} \frac{\alpha_k^t}{\alpha^t}\big( L_{k}(\bm{\omega}_k^{t}) - L_{k}(\bm{\omega}^{*}) \big) \\
&\overset{\mathrm{(a)}}{\leq} 2L\tau(\gamma^t)^2 \sum_{k=1}^{{K}} \bigg(\frac{\alpha_k^t}{\alpha^t}\bigg)^2\\
&-2\gamma^t~\sum_{k=1}^{{K}} \frac{\alpha_k^t}{\alpha^t}\big[ 1- L \gamma^t \frac{\alpha_k^t}{\alpha^t}\big]\big( L_{k}(\bm{\omega}_k^{t}) - L_{k}(\bm{\omega}^{*}) \big) \\
&\overset{\mathrm{(b)}}{\leq} 2\tau\gamma^t \bigg[1-L\gamma^t \sum_{k=1}^{{K}} \bigg(\frac{\alpha_k^t}{\alpha^t}\bigg)^2 \bigg],
\end{split}
\end{equation}
where~(\emph{a})~comes from Lemma~$1$ and ~(\emph{b})~comes from the fact that $L_{k}(\bm{\omega}_k^{t}) - L_{k}(\bm{\omega}_k^{*})\geq 0$ and Lemma $1$. Now, $\mathbb{E}[A_2]=(\gamma^t)^2 \sum_{k=1}^{{K}} \bigg(\frac{\alpha_k^t}{\alpha^t}\bigg)^2\rho_k^2$ can be found easily by applying Lemma~$1$.
Substituting $\mathbb{E}[A_1],~\mathbb{E}[A_2],~\mathbb{E}[A_3]$ into $\mathbb{E}\big(||\bm{\omega}^{t+1} - \bm{\omega}^t ||_2^2 \big)$ and using the fact that $\frac{1}{\kappa}\leq \gamma^t$ complete the proof.\\

\noindent 
\textbf{Proof of Theorem $1$}
Similar to~\cite{li2019convergence,yan2022performance}, we define $\Delta^t = \mathbb{E}[||\bm{\omega}^t - \bm{\omega}^* ||_2^2]$. From Lemma~$3$, it follows that, $\Delta^{t+1} \leq (1-\beta \gamma^t)\Delta^t + (\gamma^t)^2 G^t$. we assume $\gamma^t = \frac{\alpha}{t+\mu}$ for some $\alpha>\frac{1}{\beta}$ and $\mu>1$. Assuming $\lambda = \max\{ \frac{\alpha^2 G}{\alpha \beta -1},\mu \Delta^0\}$, we will prove $\Delta^t \leq \frac{\lambda}{t+\mu}$ by induction as follows.
The definition of $\lambda$ ensures that the inequality $\Delta^t \leq \frac{\lambda}{t+\mu}$ holds for $t=0$.
For the inequality to hold for $t>0$, it follows from definition as follows:\\ 
$\Delta^{t+1} \leq (1-\beta \gamma^t)\Delta^t + (\gamma^t)^2 G^t$

\begin{equation}
\begin{split}
&\leq \bigg(1-\frac{\alpha \beta}{t+\mu}\bigg) + \frac{\alpha^2 G^t}{(t+\mu)^2} \\
&\leq \frac{t+\mu-1}{(t+\mu)^2}\lambda + \underbrace{\bigg[ \frac{\alpha^2 G^t - \alpha \beta+1}{(t+\mu)^2}\bigg]}_{\leq 0}\\
 &\leq \frac{t+\mu-1}{(t+\mu)^2-1}\lambda \leq \frac{\lambda}{(t+1)+\mu}.
    \end{split}
\end{equation}

Specifically, if we choose $\alpha =\frac{2}{\beta}$, $\mu=\frac{2L}{\beta}$, then $\gamma^t =\frac{2}{\beta t+2L}$.
Then, we have \\
 ~~~$\lambda = \max\big\{ \frac{\alpha^2 G}{\alpha \beta -1},\mu \Delta^0\big\}$ 
 \begin{equation}
\begin{split}
&\leq \frac{\alpha^2 G}{\alpha \beta -1} +\mu \Delta^0 = \frac{4G}{\beta^2} + \frac{2L}{\beta}||\bm{\omega}^0 - \bm{\omega}^*||_2^2.\\
\end{split}
\label{eqn:lambda}
\end{equation}

\noindent 
Finally, we have: 
\\
$\mathbb{E}[L(\bm{\omega}^t)] - L^* $
 \begin{equation}
\begin{split}
&\overset{\mathrm{(a)}}{\leq} \frac{L}{2}||\bm{\omega}^t - \bm{\omega}^* ||_2^2 = \frac{L}{2} \Delta^t = \frac{L}{2} \frac{\lambda}{t+\mu}\\
&\overset{\mathrm{(b)}}{\leq} \frac{L}{2\alpha} \frac{\alpha}{t+\mu} \bigg[ \frac{4G}{\beta^2} + \frac{2L}{\beta}||\bm{\omega}^0 - \bm{\omega}^*||_2^2 \bigg]\\
&\overset{\mathrm{(c)}}{\leq}\frac{L}{\beta t +2L}\bigg[ \frac{2G}{\beta} + L ||\bm{\omega}^0 - \bm{\omega}^*||_2^2 \bigg],
\end{split}
\end{equation}
where \emph{(a)} comes from $L$-smoothness of the loss function and using the fact that $\nabla L(\bm{\omega}^*) =0$, \emph{(b)} is computed using ~Eq.~\eqref{eqn:lambda}, and \emph{(c)} is computed by substituting the values of $\alpha$ and $\mu$. Hence, the convergence is proved to be~$\mathcal{O}(\frac{1}{T})$.

\vspace{-.1cm}
\section{Acknowledgement}
The material is based upon work supported by NASA under award No(s) 80NSSC20M0261, and NSF grants 1948511, 1955561, and 2212565. Any opinions, findings, and conclusions or recommendations expressed in this material are those of the author(s) and do not necessarily reflect the views of the National Aeronautics and Space Administration (NASA) and the National Science Foundation (NSF).

\vspace{-.1cm}

\bibliographystyle{IEEEtran}
\bibliography{journalbibl}
\end{document}